\documentclass[10pt,twocolumn,letterpaper]{article}
\usepackage{titling}

\usepackage[pagenumbers]{cvpr} 

\usepackage{graphicx}
\usepackage{multirow}
\usepackage{amsmath}
\usepackage{amssymb}
\usepackage{booktabs}
\usepackage{xcolor}
\usepackage[accsupp]{axessibility}  

\usepackage{cuted}
\usepackage{bm}
\usepackage{tabulary}
\newcolumntype{x}[1]{>{\centering\arraybackslash}p{#1pt}}

\newlength\savewidth

\usepackage[pagebackref,breaklinks,colorlinks]{hyperref}

\usepackage[titletoc,title]{appendix}

\usepackage[capitalize]{cleveref}
\crefname{section}{Sec.}{Secs.}
\Crefname{section}{Section}{Sections}
\Crefname{table}{Table}{Tables}
\crefname{table}{Tab.}{Tabs.}

\newcommand{\tablestyle}[2]{\setlength{\tabcolsep}{#1}\renewcommand{\arraystretch}{#2}\centering\footnotesize}

\def\cvprPaperID{2972} 
\def\confName{CVPR}
\def\confYear{2022}

\begin{document}

\title{\vspace{-1.5ex}Affordance Diffusion: Synthesizing Hand-Object Interactions \vspace{-1.7ex}}
\author{
Yufei Ye\textsuperscript{1}\thanks{Yufei was an intern at NVIDIA during the project.} \qquad Xueting Li\textsuperscript{2} \qquad Abhinav Gupta\textsuperscript{1} \qquad Shalini De Mello\textsuperscript{2} \\ \qquad Stan Birchfield\textsuperscript{2} \qquad Jiaming  Song\textsuperscript{2}  \qquad Shubham Tulsiani\textsuperscript{1} \qquad Sifei Liu\textsuperscript{2}   \\
\textsuperscript{1}Carnegie Mellon University  \qquad \textsuperscript{2}NVIDIA \\
{\tt \small \href{https://judyye.github.io/affordiffusion-www}{https://judyye.github.io/affordiffusion-www}}
}


\maketitle

\begin{strip}\centering
\vspace{-2.1cm}
\includegraphics[width=\textwidth]{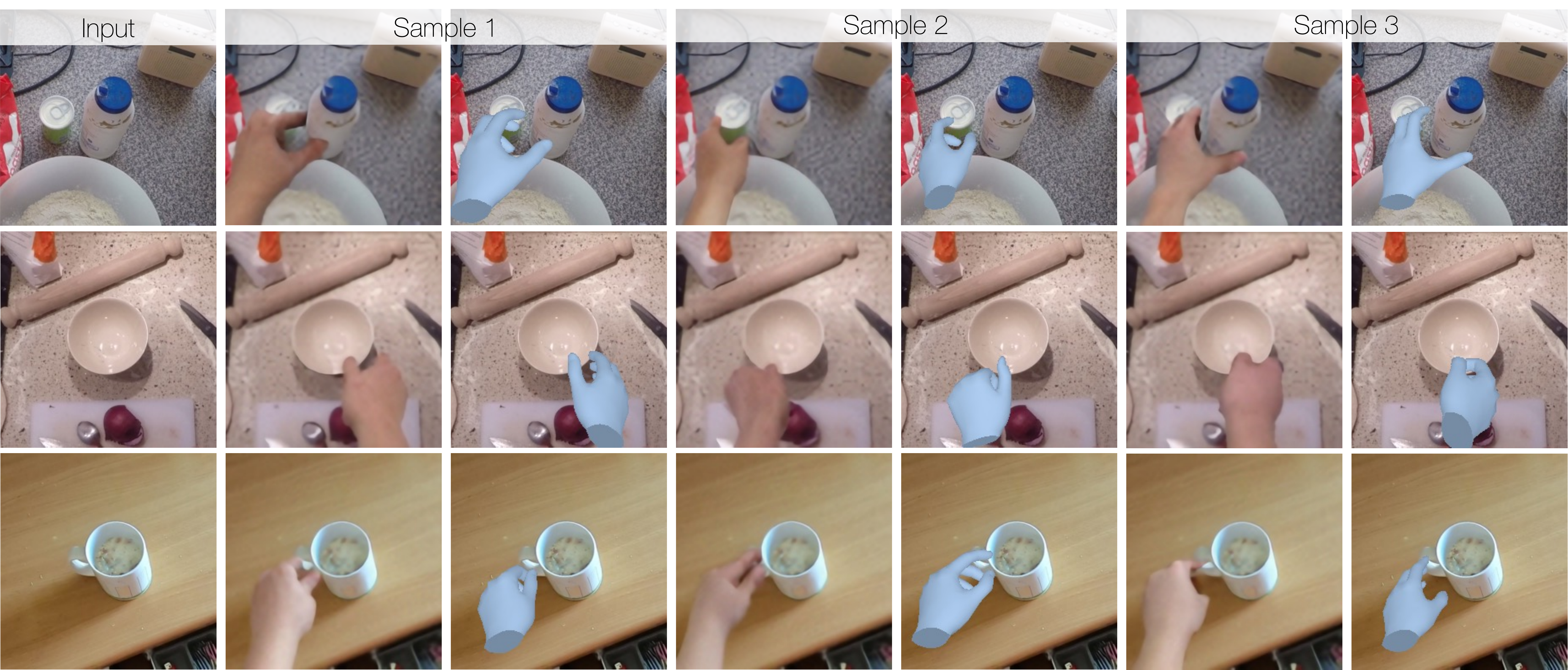}
\vspace{-0.7cm}
\captionof{figure}{Given a single RGB image of an object (first column), we synthesize plausible images of hand-object interactions from which feasible 3D hand poses can be directly extracted (remaining columns).
\label{fig:teaser}}
\vspace{-.2cm}
\end{strip}

\begin{abstract}
\vspace{-3mm}
Recent successes in image synthesis are powered by large-scale diffusion models.  However, most methods are currently limited to either text- or image-conditioned generation for synthesizing an entire image, texture transfer or inserting objects into a user-specified region. In contrast, in this work we focus on synthesizing complex interactions (\ie, an articulated hand) with a given object.  Given an RGB image of an object, we aim to hallucinate plausible images of a human hand interacting with it. 
We propose a two-step generative approach:  a LayoutNet that samples an articulation-agnostic hand-object-interaction layout, and a ContentNet that synthesizes images of a hand grasping the object given the predicted layout. Both are built on top of a large-scale pretrained diffusion model to make use of its latent representation. 
Compared to baselines, the proposed method is shown to generalize better to novel objects and perform surprisingly well on out-of-distribution in-the-wild scenes of portable-sized objects. The resulting system allows us to predict descriptive affordance information, such as hand articulation and approaching orientation.

\end{abstract}

\section{Introduction}
\label{sec:intro}
Consider the bottles, bowls and cups shown in the left column of Figure~\ref{fig:teaser}. How might a human hand interact with such objects?  Not only is it easy to imagine, from a single image, the types of interactions that might occur (\eg, `grab/hold'), and the interaction locations that might happen (\eg `handle/body'), but it is also quite natural to hallucinate---in vivid detail--- several ways in which a hand might contact and use the objects. This ability to predict and hallucinate hand-object-interactions (HOI) is critical to functional understanding of a scene, as well as to visual imitation and manipulation. 

Can current computer vision algorithms do the same?
On the one hand, there has been a lot of progress in image generation, such as synthesizing realistic high-resolution images spanning a wide range of object categories~\cite{sauer2022stylegan,kumari2022ensembling} from human faces to ImageNet classes. Newer diffusion models such as Dall-E~2~\cite{ramesh2022hierarchical} and Stable Diffusion~\cite{rombach2022high} can generate remarkably novel images in diverse styles. In fact, highly-realistic HOI images can be synthesized from simple text inputs such as ``a hand holding a cup''~\cite{ramesh2022hierarchical,rombach2022high}. 

On the other hand, however, such models fail when conditioned on an image of a particular object instance. 
Given an image of an object, it remains an extremely challenging problem to generate realistic human object interaction. Solving this problem requires (at least implicitly) an understanding of physical constraints such as collision and force stability, as well as modeling the semantics and functionality of objects --- the underlying affordances \cite{gibson1978ecological}. For example, the hand should prefer to grab the kettle handle but avoid grabbing the knife blade. Furthermore, in order to produce visually plausible results, it also requires modeling occlusions between hands and objects, their scale, lighting, texture, \etc.

In this work, we propose a method for interaction synthesis that addresses these issues using diffusion models. In contrast to a generic image-conditioned diffusion model, we build upon the classic idea of disentangling \emph{where} to interact (\emph{layout}) from \emph{how} to interact (\emph{content})~\cite{hermans2011affordance,gupta20113d}. Our key insight is that diverse interactions largely arise from hand-object layout, whereas hand articulations are driven by local object geometry.  For example, a mug can be grasped by either its handle or body, but once the grasping location is determined, the placement of the fingers depends on the object's local surface and the articulation will exhibit only subtle differences.  We operationalize this idea by proposing a two-step stochastic procedure: 
1) a \emph{LayoutNet} that generates 2D spatial arrangements of hands and objects, and
2) a \emph{ContentNet} that is conditioned on the query object image and the sampled HOI layout to synthesize the images of hand-object interactions. These two modules are both implemented as image-conditioned diffusion models.

We evaluate our method on HOI4D and EPIC-KITCHEN~\cite{liu2022hoi4d,epic}. Our method outperforms generic image generation baselines, and the extracted hand poses from our HOI synthesis are favored in user studies against baselines that are trained to directly predict hand poses. We also demonstrate surprisingly robust generalization ability across datasets, and we show that our model can quickly adapt to new hand-object-interactions with only a few examples. Lastly, we show that our proposed method enables editing and guided generation from partially specified layout parameters. This allows us to reuse heatmap prediction from prior work~\cite{fang2018demo2vec,nagarajan2019iccv:hotspots} and to generate consistent hand sizes for different objects  in one scene.


Our main contributions are summarized below: 1) we propose a two-step method to synthesize hand-object interactions from an object image, which allows affordance information extracted from it; 2) we use inpainting techinuqes to supervise the model with paired real-world HOI and object-only images and propose a novel data augmentation method to alleviate overfit to artifacts; and 3) we show that our approach generates realistic HOI images along with plausible 3D poses and generalizes surprisingly well on out-of-distribution scenes. 4) We also highlight several applications that would benefit from such a method.

\section{Related Work}
\label{sec:related}
\noindent\textbf{Understanding Hand-Object-Interaction. } In order to understand hand-object-interaction, efforts have been made to locate the active objects and hands in contact in 2D space, via either  bounding boxes detection  \cite{shan2020understanding, bambach2015lending,mittal2011hand } or segmentation \cite{shan2021cohesiv,florence2020robot}. Many works reconstruct the underlying shape of hands and objects from RGB(D) images or videos by either template-based \cite{hamer2010object,brahmbhatt2020contactpose,tekin2019h+,sridhar2016real,garcia2018first} or template-free methods \cite{hasson2019learning,karunratanakul2020grasping,ye2022hand,corona2022lisa}. Furthermore, temporal understanding of HOI videos \cite{hamer2009tracking,girdhar2021anticipative,purushwalkam2020aligning,price2022unweavenet,spriggs2009temporal} aims to locate the key frames of state changes and time of contact. In our work, we use these techniques to extract frames of interests for data collection and to analyze the synthesis results. While these works recognize what is going on with the underlying hands and objects, our task is  to hallucinate what hands could possibly do with a given object.



\noindent\textbf{Visual Affordance from Images. } Affordance is defined as functions that environments could offer \cite{gibson1978ecological}. Although the idea of functional understanding is core to visual understanding, it is not obvious what is the proper representation for object affordances. Some approaches directly map images to categories, like holdable, pushable, liftable, \etc \cite{cai2016understanding,hermans2011affordance, nagarajan2020ego,lee2015predicting}.  Some other approaches ground these action labels to images by predicting heatmaps that indicate interaction possibilities \cite{nagarajan2019iccv:hotspots, liu2022joint, fang2018demo2vec,huang2018predicting,pan2017salgan}. While heatmaps only specify \textit{where} to interact without telling \textit{what} to do, recent approaches predict richer properties such as contact distance \cite{karunratanakul2020grasping}, action trajectory\cite{mo2021iccv:wheretoact,liu2022joint}, grasping categories\cite{goyal2022human,mandikal2022dexvip}, \etc.  
Instead of predicting  more sophisticated interaction states, we
explore directly synthesizing HOI images for possible interactions because images demonstrate both \textit{where} and \textit{how} to interact comprehensively and in a straightforward manner.


\begin{figure*}
    \centering
    \includegraphics[width=\linewidth]{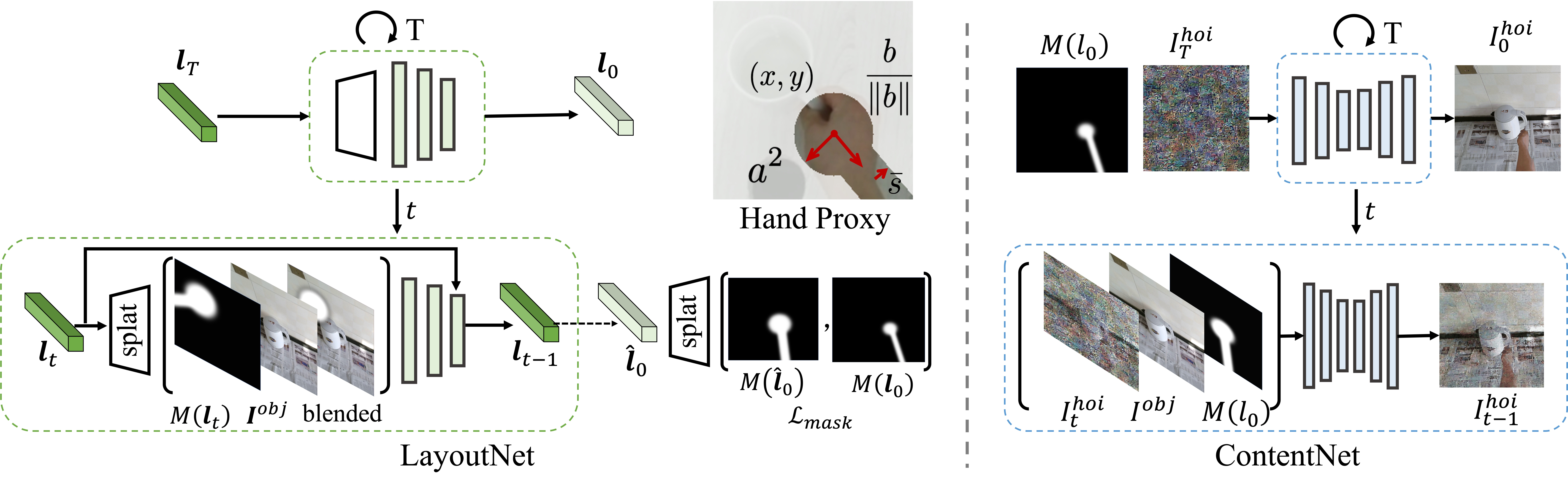}
\vspace{-0.8cm}
    \caption{The proposed method consists of two image-conditioned diffusion models: LayoutNet and ContentNet. Given an object image, we first use LayoutNet (left) to predict a HOI spatial arrangement $\bm l_0$. For every diffusion step, the LayoutNet splats the noisy layout parameter into image space, concatenates it with the object image and their blending, and predicts the denoised layout. We apply the diffusion loss in the splatted 2D space 
    $\mathcal{L}_{\text{mask}}$. 
    Then the ContentNet (right) takes in the predicted layout along with the object image to synthesize an HOI image. The two modules are connected by the articulation-agnostic hand proxy (middle top). 
    }
    \label{fig:pipeline}
\vspace{-0.3cm}
\end{figure*}

\noindent\textbf{3D Affordance. } While 3D prediction from images is common for human-scene interaction \cite{grabner2011makes,gupta20113d,fouhey2012people,li2019putting,PLACE:3DV:2020,PSI:2019}, 3D affordance for hand and object typically takes in \textit{known} object 3D shapes and predicts \textit{stable} grasps \cite{miller2004graspit,jiang2021hand,grady2021contactopt,pas2015using}. In contrast, our work predicts \textit{coarse} 3D grasps from RGB images of \textit{general} objects. The coarse but generalizable 3D hand pose prediction is shown as useful human prior for dexterous manipulation \cite{kokic2020learning,mandikal2022dexvip,dasari2022pgdm,wu2022learning,qin2022dexmv,bahl2022human}. 
While some recent works \cite{karunratanakul2020grasping,ganhand} generate 3D hand poses from images by direct regression, we instead first synthesize HOI images from which 3D hand poses are reconstructed afterwards.


\noindent\textbf{Diffusion Models and Image Editing. } Diffusion models~\cite{ho2020denoising,song2020denoising} have driven significant advances in various domains~\cite{zeng2022lion,zhou20213d,tashiro2021csdi,xu2022geodiff,kong2020diffwave}, including image synthesis~\cite{ramesh2022hierarchical,saharia2022photorealistic,balaji2022ediffi,rombach2022high}. A key advantage of diffusion models over other families of generative models~\cite{kingma2013auto,goodfellow2020generative} is their ability to easily adapt to image editing and re-synthesis tasks without much training~\cite{gal2022image,ruiz2022dreambooth,kawar2022imagic}. 
While recent image-conditioned generative models achieve impressive results on various image translation tasks such as image editing \cite{meng2021sdedit,text2live,pix2pix}, style transfer\cite{li2022diffusion,patashnik2021styleclip}, the edits mostly modify textures and style, but preserve structures, or insert new content to user-specified regions
\cite{avrahami2022blended,saharia2022palette,ramesh2022hierarchical}. In contrast, we focus on affordance synthesis where both layout (structure) and appearance are automatically reasoned about.

\section{Method}
\label{sec:method}

\def\rvx{{\mathbf{x}}}
\def\rvc{{\mathbf{c}}}
\def\rmI{{\mathbf{I}}}

Given an image of an object, we aim to synthesize images depicting plausible ways of a human hand interacting with it. Our key insight is that this multi-modal process follows a coarse-to-fine procedure. For example, a mug can either be held by its handle or body, but once decided, the hand articulation is largely driven by the local geometry of the mug. We operationalize this idea by 
proposing a two-step stochastic approach as shown in Fig ~\ref{fig:pipeline}.  

Given an object image, we first use a LayoutNet  to predict plausible spatial arrangement of the object and the hand (Sec~\ref{sec:layout}). The LayoutNet predicts hand proxy that abstracts away appearance and explicitly specifies  2D location, size and approaching direction of a grasp. This abstraction allows global reasoning of hand-object relations and also enables users to specify the interactions. Then, given  the predicted hand proxy and the object image, we synthesize a plausible appearance of an HOI via a ContentNet (Sec~\ref{sec:appearance}).  This allows the network to implicitly reason about  3D wrist orientation, finger placement, and occlusion based on the object's local shape. We use conditional diffusion models for both networks to achieve high-quality layout and visual content.
The synthesized HOI image is realistic such that a feasible 3D hand pose can be directly extracted from it by an off-the-shelf hand pose reconstruction model (Sec \ref{sec:3d}).


To supervise the system, we need pixel-aligned pairs of HOI images and object-only images that depict the exact same objects from the exact same viewpoints with the exact same lighting.  We obtain such pairs by inpainting techniques that remove humans from HOI images. We further propose a novel data augmentation to prevent the trained model from overfitting to the inpainting artifacts (Sec~\ref{sec:data}).

\subsection{Preliminary: Diffusion models}
Diffusion models are probabilistic models~\cite{sohl-dickstein2015deep,ho2020denoising} that learn to generate samples from a data distribution $p(\rvx)$ by sequentially transforming samples from a tractable distribution $p(\rvx_T)$ (\eg, Gaussian distribution).
There are two processes in diffusion models: 1) a forward noise process $q(\rvx_t|\rvx_{t-1})$ that gradually adds a small amount of noise and degrades clean data samples towards the prior Gaussian distribution; 2) a learnable backward denoising process $p(\rvx_{t-1} | \rvx_t)$ that is trained to remove the added noise. The backward process is implemented as a neural network. During inference, a noise vector $\rvx_T$ is sampled from the Gaussian prior and is sequentially denoised by the learned backward model~\cite{song2020denoising,song2020score}. 
The training of a diffusion model can be treated as training a denoising autoencoder for L2 loss~\cite{vincent2011a} at various noise levels, \textit{i.e.}, denoise $\rvx_0$ for different $\rvx_t$ given $t$. We adopt the widely used loss term in Denoising Diffusion Probabilistic Models (DDPM)~\cite{ho2020denoising,song2020denoising}, which reconstructs the added noise that corrupted the input samples. Specifically, we use the notation $\mathcal{L}_{\text{DDPM}}[\rvx; \rvc]$ to denote a DDPM loss term that performs diffusion over $\rvx$ but is also conditioned on $\rvc$ (that are not diffused or denoised): 
\begin{align}
\mathcal L_{\text{DDPM}}[\rvx; \rvc] = \mathbb E _{(\rvx, \rvc), \epsilon \sim \mathcal N(0, I), t} \|\rvx - D_\theta(\rvx_t, t, \rvc) \|_2^2,
\label{eq:ddpm}
\end{align}
where $\rvx_t$ is a linear combination of the data $\rvx$ and noise $\epsilon$, and $D_\theta$ is a denoiser model that takes in the noisy data $\rvx_t$, time $t$ and condition $\rvc$. This also covers the unconditional case as we can simply set $\rvc$ as some null token like $\varnothing$~\cite{ho2022classifier}. 

\subsection{LayoutNet: predicting where to grasp }
\label{sec:layout}

Given an object image $\rmI^{obj}$, the LayoutNet aims to generate a plausible HOI layout $\bm l$ from the learned distribution $p(\bm l | \rmI^{obj})$. 
We follow the diffusion model regime that sequentially denoises a noisy layout parameter to output the final layout. For every denoising step, the LayoutNet takes in the (noisy) layout parameter along with the object image and denoises it sequentially, \ie $\bm{l}_{t-1} \sim  \phi(\bm{l}_{t-1}|\bm l_{t}, \rmI^{obj}) $. We splat the layout parameter onto the image space to better reason about 2D spatial relationships to the object image and we further introduce an auxiliary loss term to train diffusion models in the layout parameter space. 

\noindent\textbf{Layout parameterization.}  Hands in HOI images typically appear as hands (from wrist to fingers) with forearms. Based on this observation, we introduce an articulation-agnostic hand proxy that only preserves this basic hand structure. As shown in Fig~\ref{fig:pipeline}, the layout parameter consists of hand palm size $a^2$, location $x,y$ and approaching direction $\arctan(b_1, b_2)$, \ie $\bm l:= (a, x, y, b_1, b_2)$. The ratio of hand palm size and forearm width $\bar s$ remains a constant that is set to the mean value over the training set.
We obtain the ground truth parameters from hand detection (for location and size) and hand/forearm segmentation (for orientation).



\noindent\textbf{Predicting Layout.} The diffusion-based LayoutNet takes in a noisy 5-parameter  vector $\bm{l}_t$ with the object image and outputs the denoised layout vector $\bm{l}_{t-1}$ (we define $l_0 = l$). To better reason about the spatial relation between hand and object, we splat the layout parameter into  the image space $M(\bm{l}_t)$. The splatted layout mask is then concatenated with the object image and is passed to the diffusion-based LayoutNet.
We splat the layout parameter to 2D by the spatial transformer network~\cite{jaderberg2015spatial} that transforms a canonical mask template by a similarity transformation.  




\noindent\textbf{DDPM loss for layout.} 
One could directly train the LayoutNet with the DDPM loss (Eq.~\ref{eq:ddpm}) in the layout parameter space: $\mathcal L_{para} := \mathcal L_{\text{DDPM}}[\bm{l}; \rmI^{obj}]$. 
However, when diffusing in such a space, multiple parameters can induce an identical layout, such as a size parameter with opposite signs or approaching directions that are scaled by a constant.  
DDPM loss in the parameter space would penalize predictions even if they guide the parameter to a equivalent one that induce the same layout masks as the ground truth. 
As the downstream ContentNet only takes in the splatted masks and not their parameters, we propose to directly apply the DDPM loss in the splatted image space (see appendix for details): 
\begin{align}
\mathcal L_{mask} = \mathbb E _{(\bm{l}_0, \rmI^{obj}), \epsilon \sim \mathcal N(0, I), t} \|M(\bm{l}_0) - M(\hat {\bm l}_0) \|_2^2.
\end{align}
where $\hat {\bm{l}}_0 := D_\theta(\bm{l}_t, t, \rmI^{obj})$ is the output of our trained denoiser that takes in the current noisy layout $\bm{l}_t$, the time $t$ and the object image $\rmI^{obj}$ for conditioning. 

In practice, we apply losses in both the parameter space and image spaces $\mathcal L_{mask} + \lambda \mathcal L_{para}$ 
because when the layout parameters are very noisy in the early diffusion steps, the splatted loss in 2D alone is a too-weak training signal.  

\noindent\textbf{Network architecture.} We implement the backbone network as a UNet with cross-attention layers and initialize it from the pretrained diffusion model~\cite{nichol2021glide}. The model takes in images with seven channels as shown in Fig~\ref{fig:pipeline}: 3 for the object image, 1 for the splatted layout mask and another 3 that blends the layout mask with object image.  The noisy layout parameter attends spatially to the feature grid from the UNet's bottleneck  and spit out the denoised output.

\noindent\textbf{Guided layout generation. } The LayoutNet is trained to be conditioned on an object image only but the generation can be guided with  additional conditions at test time without retraining. For example, we can condition the network to generate layouts such that their locations are at certain places \ie $\bm l\sim p(\bm l_0|\rmI^{obj},x=x_0, y=y_0)$. We use techniques~\cite{song2020score} in diffusion models that hijack the conditions after each diffusion steps with corresponding noise levels. This guided diffusion enables user editing and HOI synthesis for scenes with a consistent hand scale (Sec.~\ref{sec:app}). Please refer to the appendix for LayoutNet implementation details. 


\subsection{ContentNet: predicting how to grasp}
\label{sec:appearance}
Given the sampled layout $\bm l$ and the object image $\rmI^{obj}$, the ContentNet synthesizes a HOI image $\rmI^{hoi}$. While the synthesized HOI images should respect the provided layout, the generation is still stochastic because hand appearance may vary in shape, finger articulation, skin colors, \etc.  
We leverage the recent success of diffusion models in image synthesis and formulate the articulation network as a image-conditioned diffusion model.  
As shown in Fig~\ref{fig:pipeline}, at each step of diffusion, the network takes in channel-wise concatenation of the noisy HOI image, the object image and the splatted mask from the layout parameter and outputs the denoised HOI images $D_\phi(\rmI^{hoi}_t, t, [\rmI^{obj}, M(\bm l)])$. 


We implement the image-conditioned diffusion model in the latent space~\cite{ldm,vahdat2021score,sinha2021d2c} and finetune it from the inpainting model that is pre-trained on large-scale data. The pretraining is beneficial as the model has learned the prior of retaining the pixels in unmask region and hallucinate to fill the masked region. During finetuning, the model further learns to respect the predicted layout, \ie, retaining the object appearance if not occluded by hand and synthesizing hand and forearm appearance depicting finger articulation, wrist orientation, etc.

\begin{figure}
    \centering
    \includegraphics[width=0.85\linewidth]{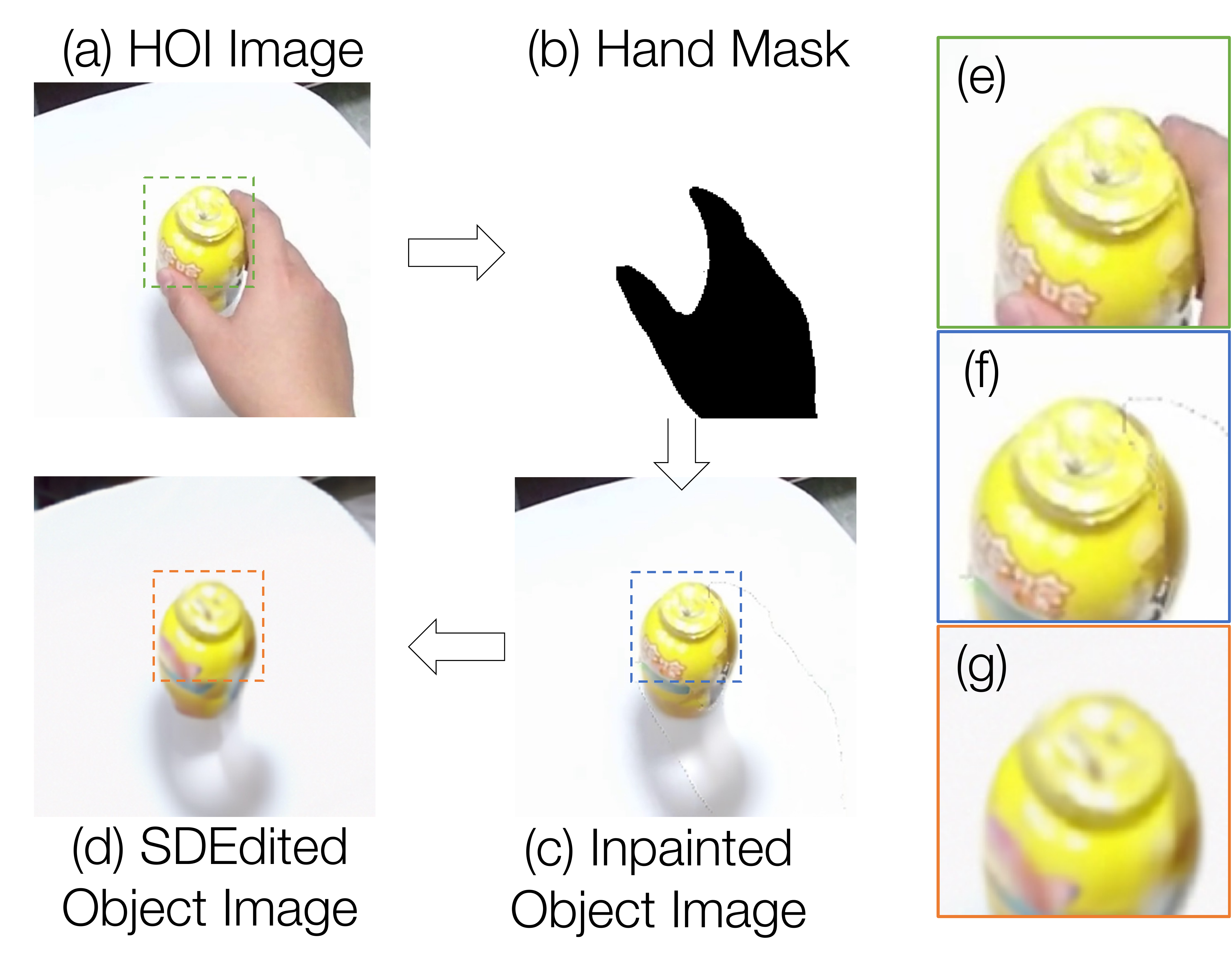}
    \vspace{-0.5em}    
    \caption{\textbf{Paired Data Generation:} Given an HOI  image, we first segment out hand (b) and remove it by inpainting (c). Then we use SDEdit~\cite{sdedit} to reduce inpainting artifact (d). As inpainting introduce discrepancy between mask and unmasked region (f) while SDEdit undesirably modifies the unmasked object region, we mix up \textit{both} object image sets in training. }
    \label{fig:data}
    \vspace{-1.5em}
\end{figure}
\subsection{Constructing Paired Training Data}
\label{sec:data}


To train such a system, we need pairs of object-only images and HOI image. These pairs need to be pixel-aligned except for the hand regions. One possible way is to use synthetic data~\cite{hasson2019learning, ganhand} and render their 3D HOI scene with and without hands. But this introduces domain gap between simulation and the real-world thus hurts generalization. We instead follow a different approach. 

As shown in Fig~\ref{fig:data}, we first extract object-centric HOI crops from egocentric videos with 80\% square padding. Then we segment the hand regions to be removed and pass them to the inpainting system~\cite{nichol2021glide} to hallucinate the objects behind hands. The inpainter is trained on millions of data with people filtered out therefore it is suitable for our task. 

\noindent\textbf{Data Augmentation.} Although the inpainting  generates impressive object-only images, it still introduces editing artifacts, which the networks can easily overfit to~\cite{zhang2020learning}, such as sharp  boundary and blurriness in masked regions. 
We use SDEdit~\cite{meng2021sdedit} to reduce the discrepancy between the masked and unmasked regions. SDEdit first adds a small amount of noise (we use $5\%$ of the whole diffusion process) to the given image and then denoises it to optimize overall image realism. However, although the discrepancy within images reduces, the unmasked object region is undesirably modified and the overall SDEdited images appear blurrier.  
In practice, we mix up the object-only images with and without SDEdit for training. 


We collect all data pairs from HOI4D~\cite{liu2022hoi4d}. After some automatic sanity filtering (such as ensuring hands are removed),  we generate 364k pairs of object-only images and HOI-images in total. We call the dataset HO3Pairs (Hand-Object interaction and Object-Only Pairs). We provide details and more examples of the dataset in the appendix. 



\section{Experiments}
\label{sec:exp}
\begin{figure*}
    \centering
    \includegraphics[width=\linewidth]{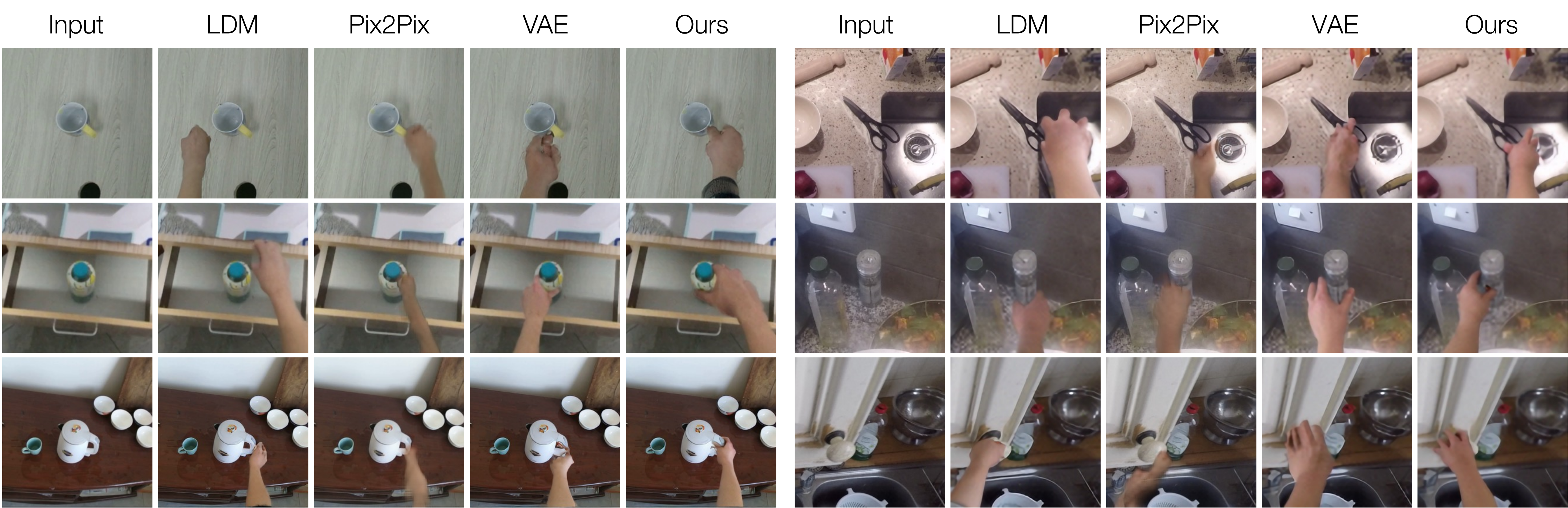}
    \vspace{-0.5cm}
    \caption{Visualizing HOI synthesis from our method and three
baselines~\cite{ldm,pix2pix,kingma2013auto} on HOI4D (left) and EPIC-KITCHEN dataset (right).}
    \label{fig:syn2d}
\vspace{-0.2cm}
 \end{figure*} 
\begin{table*}[t]
\vspace{-0.5em}
\caption{Quantitative results for HOI synthesis using contact recall, FID score, and a user study on the HOI4D and EPIC-KITCHEN datasets. We  compare our method with prior works \cite{ldm,pix2pix,kingma2013auto}.
\label{tab:syn2d}
\vspace{-0.5em}}
\centering
\tablestyle{3pt}{1}
\begin{tabular}{lcccccccccccc  ccc }
\toprule
 & \multicolumn{12}{c}{HOI4D dataset} & \multicolumn{3}{c}{EPIC-KITCHEN dataset} \\
\cmidrule(l{2pt}r{2pt}){2-13} \cmidrule(l{2pt}r{2pt}){14-16}
\multirow{2}*{Method} & \multicolumn{10}{c}{Contact Recall(\%) } & \multirow{2}*{FID} & User & Contact & \multirow{2}*{FID} & User   \\
& Kettle	&Knife	&TrashCan	&Chair	&Mug	&Bowl	&ToyCar	&Laptop	& Bottle & mean & & Study & Recall & &Study \\
\midrule
LDM   \cite{ldm}  	&82.67	&72.28	&83.33	&82.08	&66.67	&78.10	&88.00	&62.00	&87.22	& 64.44 & 105.26 &	 27.5 & 76.56 & 118.15 & 23.3 \\
Pix2Pix \cite{pix2pix} &79.50	& 70.26	& 82.50	& 76.88	& 68.50	& 79.64	& 89.00	& 63.00	& 85.42 & 73.02	& 107.09 &	 15.5 & 70.00 & 125.62 & 13.3\\
VAE-ContentNet \cite{kingma2013auto} & 91.00 &	78.95 &	91.50	&85.63	&73.00	&90.00	&94.00	&69.00	&90.00 &83.49	& \textbf{98.19}	& 23.0 & 82.03 & \textbf{115.86} & 27.9 \\ 
Ours    & 91.00	& 84.21	&97.00	&88.75	&60.00	&92.86	&96.00	& 72.00	& 91.67       & \textbf{87.14}	& 99.00	&  \textbf{34.0} & \textbf{86.56} & 117.22 & \textbf{35.4} \\
\bottomrule
\end{tabular}
\vspace{-.5cm}
\end{table*}

We train our model on the contructed HO3Pairs dataset, evaluate it on the HOI4D~\cite{liu2022hoi4d} dataset and show zero-shot generalization to the EPIC-KITCHEN~\cite{epic} dataset.  
We evaluate both the generated HOI images and the extracted 3D poses. 
For image synthesis, we compare with conditional image synthesis baselines and show that our method generates more plausible hands in interaction.
Beyond 2D HOI image synthesis, we compare the extracted 3D poses with prior works that directly predict 3D hand poses.
%
%
Furthermore, we show several applications enabled by the proposed HOI synthesis method, including few-shot adaptation, image editing by layout, heatmap-guided prediction and integrating object affordance with the scene. 

%

\noindent\textbf{Datasets}
%
Instead of testing with inpainted object images, we evaluate our model on the real object-only images cropped from the frames without hands. The goal is to prevent models from cheating by overfitting to the inpainting artifacts, as justified in the ablations below.

The HOI4D dataset is an egocentric video dataset recording humans in a lab environment interacting with various objects such as kettles, bottles, laptops, \etc. The dataset provides manual annotations of hand and object masks, action labels, object categories, instance ID, and ground truth 3D hand poses. We train and evaluate on 10 categories where full annotations are released. For each category, we hold out 5 object instances for evaluation. 
In total, we collect 126 testing images.

The EPIC-KITCHEN dataset displays more diverse and cluttered scenes. We construct our test set by randomly selecting 10 frames from each video clip. We detect and crop out objects without hands~\cite{detectron2}. In total, we collect 500 object-only images for testing.

\subsection{Evaluating Image Synthesis}
\noindent\textbf{Evaluation Metrics. }
We evaluate HOI generation using three metrics. First, we report the FID score~\cite{Seitzer2020FID,heusel2017gans}, which is widely used for image synthesis that measures the distance between two image sets. We generate 10 samples for every input and calculate FID  with 1000 HOI images extracted from the test sets. 
%
We further evaluate the physical feasibility of the generated hands by the contact recall metric --- it computes the ratio of the generated hands that are in the ``in-contact'' state by an off-the-shelf hand detector~\cite{shan2020understanding}.
We also carry out user studies to evaluate their perceptual plausibility. Specifically, we present two images from two randomly selected methods to users and ask them to select the more plausible one. We collect 200 (for HOI4D) and 240 (for EPIC-KITCHEN) answers and report the likelihood of the methods being chosen.

\noindent\textbf{Baselines. }
We compare our method with three strong image-conditional synthesis baselines. 
1) \noindent\textit{Latent Diffusion Model (LDM)}~\cite{ldm} is one of the state-of-the-art generic image generation models that is pre-trained with large-scale image data.  We condition the model on the object image and finetune it on HO3Pair dataset. This baseline jointly generates both layout and appearance with one network. 
2) \textit{Pix2Pix}~\cite{pix2pix} is commonly used for pose-conditioned human/hand synthesis~\cite{chan2019everybody,ganerated_hand}. We modify the model to condition on the generated layout masks that are predicted from our LayoutNet. 
3) \textit{VAE}~\cite{kingma2013auto} is a widely applied generative model in recent affordance literature~\cite{fouhey2012people,li2019putting,PLACE:3DV:2020}. This baseline uses a VAE with ResNet~\cite{he2016deep} as backbone to predict a layout parameter. The layout is then passed to our ContentNet to generate images. 
 
 \noindent\textbf{Results.} We visualize the generated HOI images in Fig~\ref{fig:syn2d}. Pix2Pix typically lacks detailed finger articulation. While LDM and VAE generate more realistic hand articulations than Pix2Pix,  the generated hands sometimes do not make contact with the objects. The hand appearance near the contact region is less realistic. In some cases, LDM does not add hands at all to the given object images. In contrast, our model can generate hands with more plausible articulation and the synthesized contact regions are more realistic. 
This is consistent with the quantitative results in Tab~\ref{tab:syn2d}. While we perform comparably to the baselines in terms of the FID score, we achieve the best in terms of contact recall. The user study shows that our results are favored the most. This may indicate that humans perceive interaction quality as a more important factor than general image synthesis quality.

\noindent\textbf{Generalizing to EPIC-KITCHEN. }
Although our model is trained only on the HOI4D dataset with limited scenes and relatively clean backgrounds, our model can generalize to the EPIC-KITCHEN dataset without any finetuning. In Fig~\ref{fig:syn2d}, the model also generalizes to interact with unseen categories such as scissors and cabinet handles. Tab~\ref{tab:syn2d} reports similar trends: performing best in  contact recall, comparably well in image synthesis and is favored the most by users. 

\begin{figure*}
    \centering
    \includegraphics[width=\linewidth]{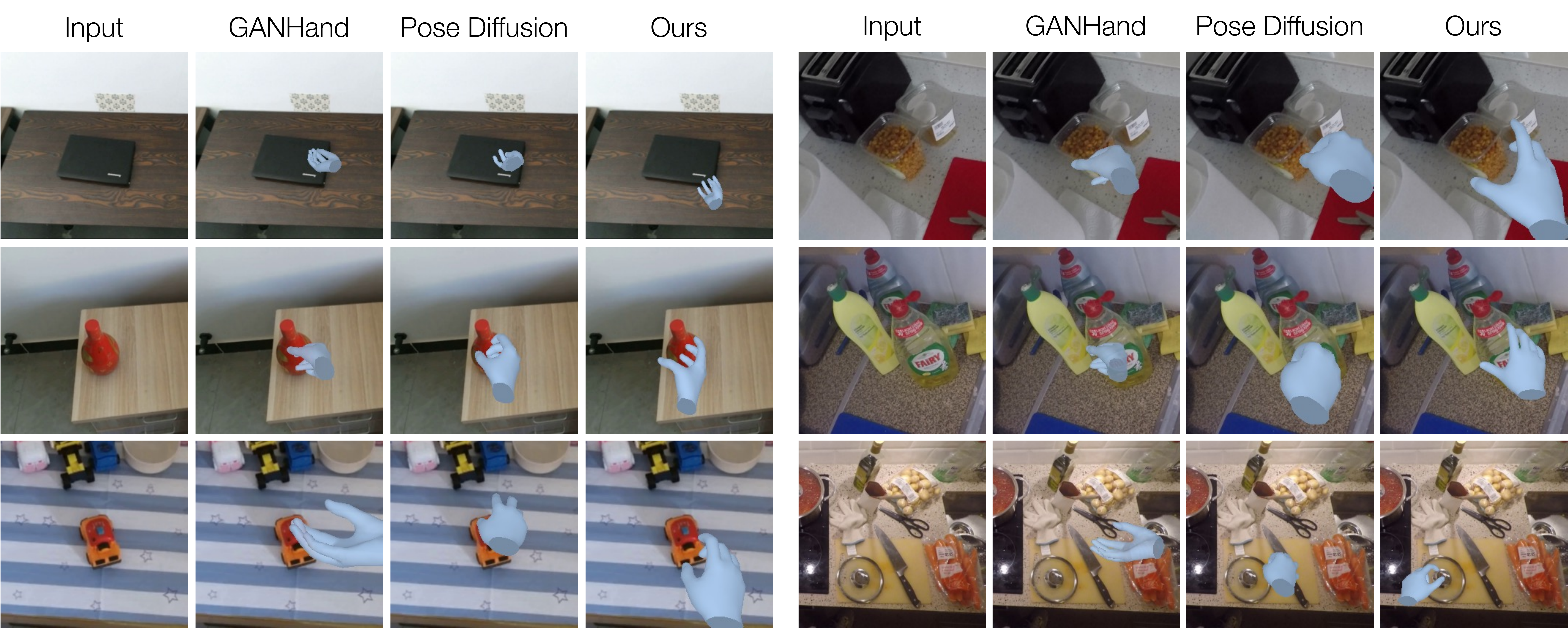}
    \caption{Visualizing 3D affordance prediction from our method,
GANHand~\cite{ganhand} and diffusion model~\cite{ldm} that directly predicts 3D pose on HOI4D (left) and EPIC-KITCHEN dataset (right).}
\vspace{-0.5cm}
    \label{fig:3d}
\end{figure*}
\begin{table}[t]
\centering
\tablestyle{2pt}{1}
\caption{\textbf{Analysis of data augmentation}: contact recall (CR\%) and FID score on the real and the inpainted object image set of HOI4D and comparisons of ours with the ablations of excluding aggressive common data augmentation (CmnAug) or SDEdit~\cite{sdedit}.}
\label{tab:artifact}
\vspace{-0.3em}
\begin{tabular}{llcccc}
\toprule
        &  & \multicolumn{2}{c}{Real Obj Img} & \multicolumn{2}{c}{Inpainted Img}   \\
\cmidrule(l{2pt}r{2pt}){3-4} \cmidrule(l{2pt}r{2pt}){5-6}         
CmnAug & SDEdit& CR & FID & CR & FID   \\
\midrule
          &            &	39.37&	113.93	&	89.05	& 89.38 \\
\checkmark &            &	79.52&	99.12	&	93.81	& 89.01 \\
\checkmark & \checkmark & 87.14  & 	99.00	&	94.29	& 88.50 \\
\bottomrule
\end{tabular}
\vspace{-0.5cm}
\end{table}
\noindent\textbf{Ablation: Data Augmentation. } Tab \ref{tab:artifact} shows the benefits of data augmentation to prevent overfitting. Without any data augmentation, the model performs well on the inpainted object images but catastrophically fails on the real ones. When we add aggressive common data augmentations like Gaussian blur and Gaussian noise, the performance improves. Training on SDEdited images further boosts the performance. The results also justify the use of real object images as test set since evaluating on the inpainted object images may not reflect the real performance.

\noindent\textbf{Ablation: LayoutNet Design. }
We analyze the benefits from our LayoutNet design by reporting contact recall. The LayoutNet predicts more physically feasible hands by taking in the splatted layout masks instead of the 5-parameter layout vector (87.14\% vs 78.10\%). Moreover, the contact recall drops to 83.96\% when the diffusion loss in Sec~\ref{sec:layout} is removed, verifying its contribution to the LayoutNet.

\subsection{Evaluating Extracted 3D Hand Poses}
\label{sec:3d}
Thanks to the realism of the generated HOI images, 3D hand poses can be directly extracted from them by an off-the-shelf hand pose estimator~\cite{frankmocap}. We conduct a user study to compare the 3D poses extracted from our HOI images against methods that directly predict 3D pose from object images. We present the rendered hand meshes overlaid on the object images to users and are asked to select the more plausible one. In total, we collected 400 and 380 answers from users for HOI4D and EPIC-KITCHEN, respectively. 


\begin{table}[t]
\centering
\tablestyle{3pt}{1}
\caption{User study for 3D affordance prediction on HOI4D and EPIC-KITCHEN dataset. We compare our method with GANHand \cite{ganhand} and a diffusion model that directly predicts 3D poses.
\label{tab:3d}
}
\begin{tabular}{l cc}
\toprule
 Method & HOI4D & EPIC  \\
\midrule
GANHand  \cite{ganhand}  & 23.8 & 23.53 \\
3D Pose Diffusion &  27.9 & 34.1 \\
Ours & \textbf{48.2} & \textbf{42.4} \\
\bottomrule
\end{tabular}
\vspace{-0.3cm}
\end{table}
\noindent\textbf{Baselines.} 
While most 3D hand pose generation works require 3D object meshes as inputs, a recent work by Corona \etal (GANHand)~\cite{ganhand} can hallucinate hand poses from an object image.
Specifically, they first map the object image to a grasp type~\cite{feix2015taxonomy} with the predefined coarse pose and then regress a refinement on top. We finetune their released model on the HO3Pairs datasets with the ground truth 3D hand poses.  We additionally implement a diffusion model baseline that sequentially diffuses 3D hand poses. The architecture is mostly the same as the LayoutNet but the diffused parameter is increased to 51 (48 for hand poses and 3 for scale and location) and the splatting function is replaced by the MANO~\cite{mano} layer that renders hand poses to image. See the appendix for  implementation details.

\noindent\textbf{Results. } As shown in Fig~\ref{fig:3d}, GANHand\cite{ganhand} predicts reasonable hand poses for some objects but fails when the grasp type is not correctly classified. The hand pose diffusion model sometimes generates infeasible hand poses like acute joint angles.  Our model is able to generate hand poses that are compatible with the objects. Furthermore, while previous methods typically assume right hands only, our model can automatically generate  both left and right hands by implicitly learning the correlation between approaching direction and hand sides.  The qualitative performance is also supported by the user study  in Tab~\ref{tab:3d}. 

\subsection{Application} 
\label{sec:app}
We showcase several applications that are enabled by the proposed method for hand-object-image synthesis. 

\begin{table}[t]
\centering
\tablestyle{3pt}{1}
\caption{\hspace{1em}\textbf{Few-shot Adaption:} Quantitative results using contact recall when finetuning the proposed HOI synthesis model and a pretrained inpainting model with 32 samples from new categories. 
}
\label{tab:fewshot}
\begin{tabular}{lcccc}
\toprule
 & bucket & scissors & stapler & mean \\
\midrule
w HOI pretrain   &92.0	&95.0	&70.0	& 85.7 \\
w/o HOI pretrain &90.0	&68.8	&34.0	& 64.3 \\
\bottomrule
\end{tabular}
\vspace{.5em}
\vspace{-.5cm}
\end{table}
\noindent\textbf{Few-shot Adaptation. } In Tab~\ref{tab:fewshot}, we show that our model can be quickly adapted to a new HOI category with as few as 32 training samples. We initialize both LayoutNet and ContentNet from our HOI4D-pretrained checkpoints and compare it with the baseline model that was pre-trained for inpainting on a large-scale image dataset~\cite{ldm}. We finetune both models on 32 samples from three novel categories in HOI4D and test with novel instances. The baseline model adapts quickly on some classes, justifying our reasons to finetune our model from them---generic large-scale image pretraining indeed already learns good priors of HOI. Furthermore, our HOI synthesis model performs even better  than the baseline.

\begin{figure}
    \centering
    \includegraphics[width=\linewidth]{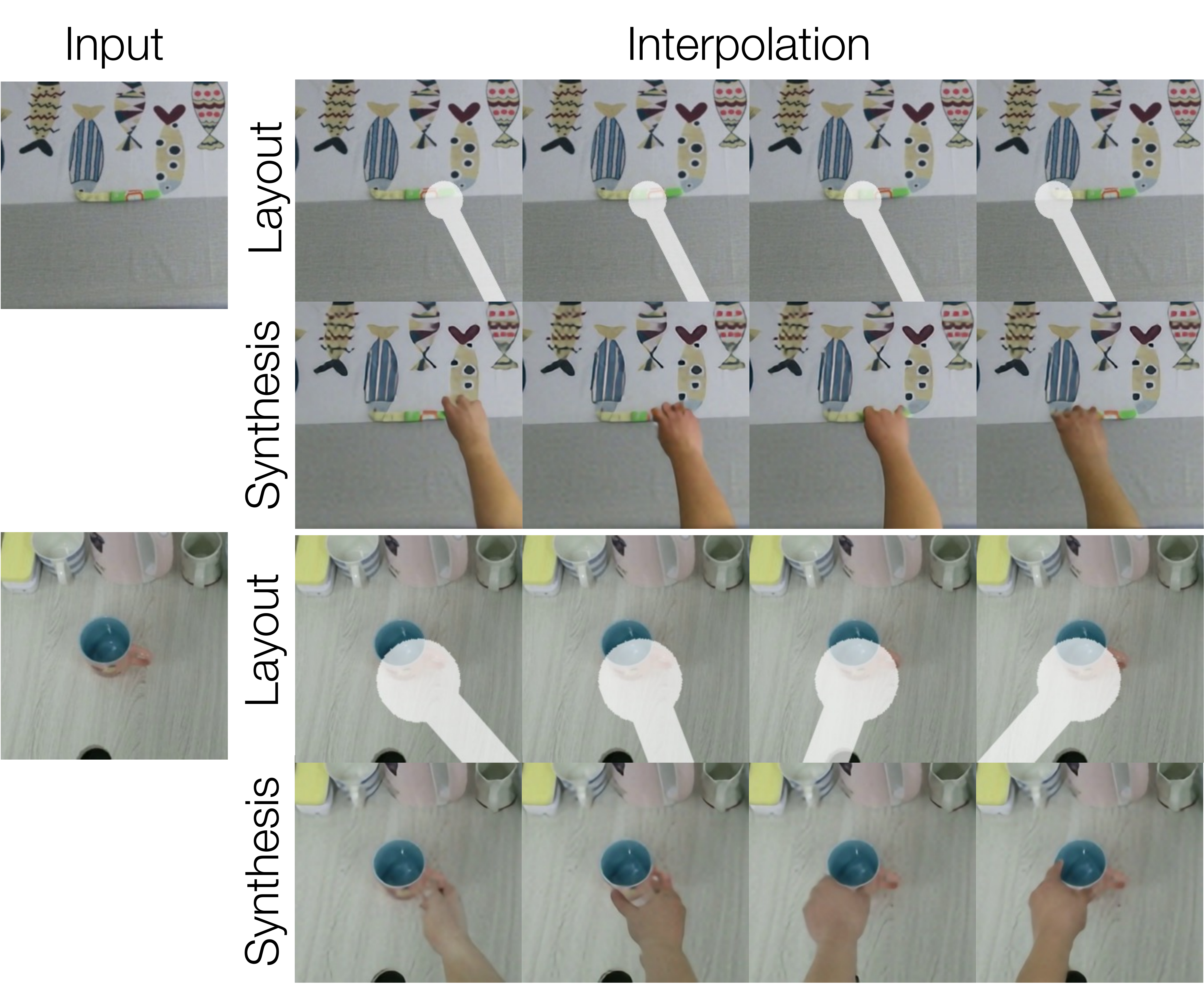}
    \caption{\textbf{Layout Editing}: Visualizing HOI synthesis when the conditioned layouts gradually change location and orientation. }
    \label{fig:interpolate}
    \vspace{-0.1cm}
\end{figure}
\noindent\textbf{Layout Editing. } The layout representation allows users to edit and control the generated hand's structure. As shown in Fig~\ref{fig:interpolate}, while we gradually change the layout's location and orientation, the synthesized hand's appearance changes accordingly. As the approaching direction to the mug changes from right to left, the synthesized fingers change accordingly from pinching around the handle to a wider grip around the mug's body.

\begin{figure}
    \centering
    \includegraphics[width=\linewidth]{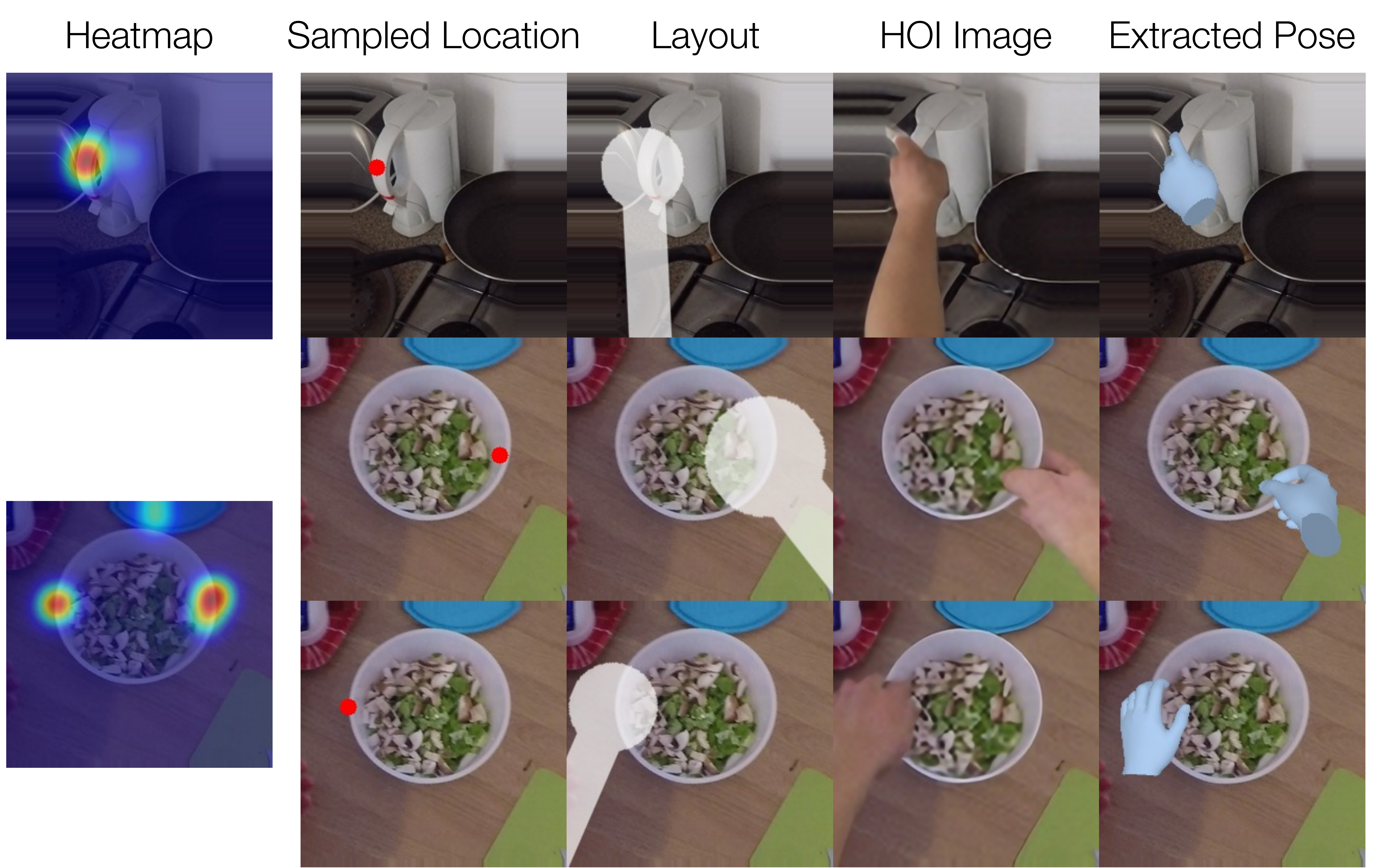}
    \caption{\textbf{Heatmap-guided synthesis: } Given a heatmap, LayoutNet is guided to generate layout at the sampled location, from which HOI images are synthesized and 3D poses are extracted.}
    \label{fig:heatmap}
    \vspace{-0.3cm}
\end{figure}
\noindent\textbf{Heatmap-Guided Synthesis.} As shown in Sec~\ref{sec:layout}, our synthesized HOI images can be conditioned on a specified location without any retraining. 
This not only allows users to edit with just keypoints, but also enables our model to utilize contact heatmap predictions from prior works~\cite{nagarajan2019iccv:hotspots,fang2018demo2vec}. In Fig~\ref{fig:heatmap}, we sample points from the heatmaps and conditionally generate layouts and HOI images which further specifies \textit{how} to interact at the sampled location.

\begin{figure}
    \centering
    \includegraphics[width=\linewidth]{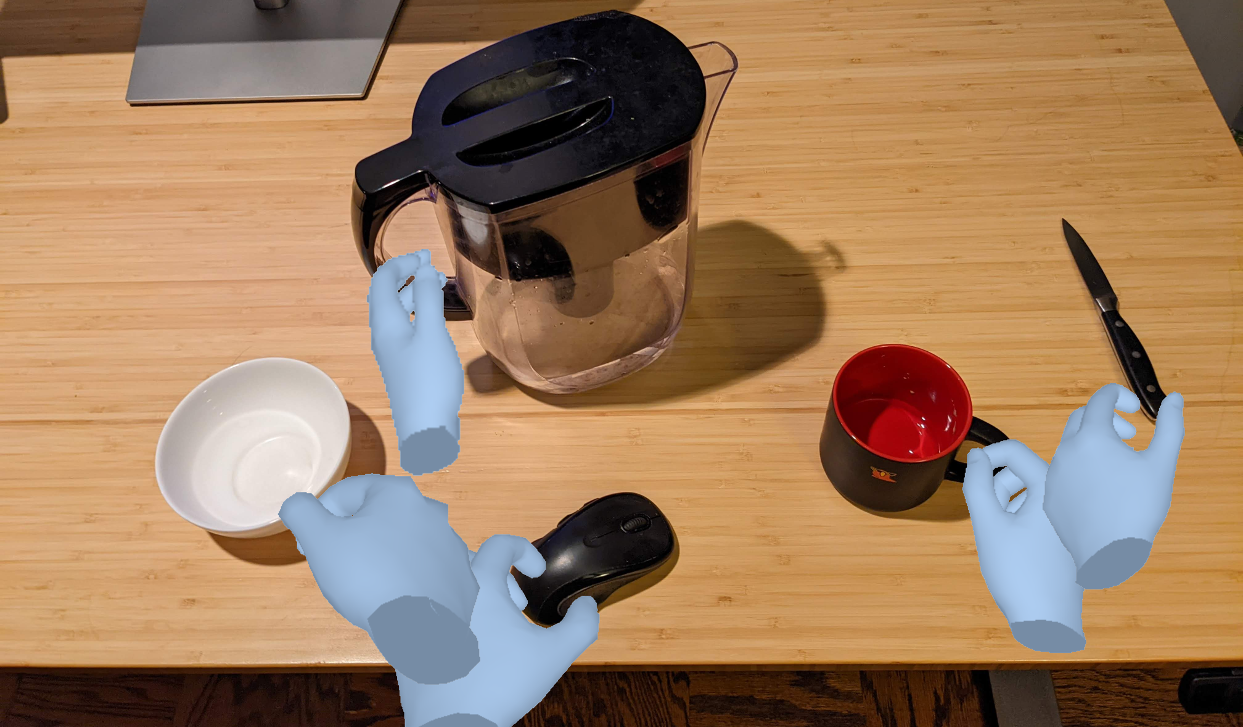}
    \caption{\textbf{Scene-level Integration: } Given a cluttered scene, we detect each object and synthesize its interactions individually. Each object's layout scale is guided to appear in the same size when transferred back to the scene.  }
    \label{fig:scene}
    \vspace{-0.5cm}
\end{figure}
\noindent\textbf{Integration to scene.} We integrate our object-centric HOI synthesis to scene-level affordance prediction. While the layout size is predicted relative to each object, hands for different objects in one scene should exhibit consistent scale. To do so, we first specify one shared hand size for each scene and calculate the corresponding relative sizes in each crops (we assume objects at similar depth and thus sizes can be transformed by crop sizes, although more comprehensive view conversions can be used). The LayoutNet is conditioned to generate these specified sizes with guided generation techniques (Sec~\ref{sec:layout}). Fig~\ref{fig:scene} shows the extracted hand meshes from each crops transferred back to the scene.

\section{Conslusion}
\label{sec:conclusion}
In this paper, we propose to synthesize hand-object interactions from a given object image. We explicitly reason about \textit{where} to interact and \textit{how} to interact by LayoutNet and ContentNet. Both of them are implemented as diffusion models to achieve controllable and high-quality visual results. The synthesized HOI images enable a shortcut to more plausible 3D affordance via reconstructing hand poses from them. Although the generation quality and the consistency between the extracted 3D poses and images can be further improved, we believe that  HOI synthesis along with our proposed solution opens doors for many promising applications and contributes towards the general goal of understanding human interactions in the wild. 

\textbf{Acknowledgements.} This work is supported by NVIDIA Graduate Fellowship
to YY. The authors would like to thank Xiaolong Wang, Sudeep Dasari, Zekun Hao and Songwei Ge for helpful discussions. 

{\small
\bibliographystyle{ieee_fullname}
\bibliography{egbib}
}

\clearpage

\title{Affordance Diffusion: Synthesizing Hand-Object Interactions\\Supplementary Materials}
\author{
Yufei Ye\textsuperscript{1*} \qquad Xueting Li\textsuperscript{2} \qquad Abhinav Gupta\textsuperscript{1} \qquad Shalini De Mello\textsuperscript{2} \\ \qquad Stan Birchfield\textsuperscript{2} \qquad Jiaming  Song\textsuperscript{2}  \qquad Shubham Tulsiani\textsuperscript{1} \qquad Sifei Liu\textsuperscript{2}   \\
\textsuperscript{1}Carnegie Mellon University  \qquad \textsuperscript{2}NVIDIA \\
}


\maketitle
\appendix


In the supplementary material, we provide more implementation details and more qualitative results. We discuss the details of articulation-agnostic hand proxy and how to apply DDPM loss in the image space for training the LayoutNet (Sec. \ref{layout}). We also present ablations on ContentNet(Sec. \ref{content}). We further show: (i) the paired data construction method being robust, in Sec. \ref{3.3}, (ii) baseline implementations details in Sec. \ref{3.4}, (iii) details of integrating our approach to scene-level affordance prediction in Sec. \ref{3.5}. Finally, we discuss the limitation of our approach (Sec. \ref{3.6}), and show more qualitative results in Sec. \ref{qr}. \textbf{Visual results are also included in the video.}

\section{Implementation Details}

\subsection{LayoutNet (Sec~3.1)}\label{layout}
\noindent\textbf{Layout parameters. }
As mentioned in Sec~3.1 of the main paper, we parameterize the layout as $(x, y, a, b_1, b_2)$, where $x,y$ is the location, $a^2$ is size, and $b_1, b_2$ are un-normalized approaching direction parameters. 
For training the LayoutNet, we obtain the ground truth parameters from off-the-shelf 2D hand prediction systems. The size and location comes from the predicted bounding box of a hand detector~\cite{shan2020understanding}, which typically defines the hand region up to the wrist. The orientation is calculated from hand segmentation whose region is typically defined as the entire hand region, including hand and forearm. The approaching direction is calculated as the first principal component of a hand mask that centers on the location of the palm of the predicted hand.

We splat the layout parameters onto 2D via the spatial transformer network~\cite{jaderberg2015spatial} that transforms a canonical mask template by a similarity transformation.  
The 2D similarity transformation is determined from the layout parameters. More formally, 

\begin{align*}
    T_l = \begin{pmatrix}sR & t \\ 0 & 1\end{pmatrix} =  \begin{pmatrix}a^2 \hat b_1 & -a^2\hat b_2 & x \\  a^2 \hat b_2 & a^2 \hat b_1 & y \\ 0  & 0 & 1\end{pmatrix},
\end{align*}
where $\hat b_1, \hat b_2 $ is the normalized vector of $b_1, b_2$. 

The lollipop-shape template in the canonical space is implemented with its circle being an isometric 2D Gaussian with a standard deviation of $1$ and its rectangle being a 1D  Gaussian with a standard deviation $\bar s = 2$. The width of the rectangle is calculated from the training data as the average ratio of the widths of forearms and palms. 

\noindent \textbf{DDPM loss on mask. }
In Eq 1 and 2 of the main paper, we write the DDPM loss in terms of reconstructing clean samples. In practice, we follow prior works~\cite{ldm,nichol2021glide,ramesh2022hierarchical} that reconstruct the added noise $\epsilon$ as
\begin{align*}
\mathcal L_{\text{DDPM}}^{\text{noise}} = \mathbb E _{x, \epsilon \sim \mathcal N(0, I), t} \|\epsilon - \epsilon_\theta(x_t, t) \|_2^2.
\end{align*}
The estimated clean sample $\hat l_0$ is connected with the estimated noise by  $ \hat l_0 = \frac{1}{\sqrt{1 - \bar \alpha_{t}}} l_{t} - \frac{\sqrt{\bar \alpha_{t}}}{\sqrt{1 - \bar \alpha_{t}}} \epsilon_\theta$, where $\alpha_t, \bar \alpha_t$ represent the noise schedule for each diffusion time step. 

We train the LayoutNet with a weighted sum of the parameter loss $\mathcal{L}_{\text{para}}$ for esitmating the noise term $\epsilon$,  and a mask loss $\mathcal{L}_{\text{mask}}$ for estimating the clean sample term $\hat l_0$. The hyperparamter $\lambda$ is set to $10$. 

\noindent\textbf{Guided layout generation. }
LayoutNet inherits properties from diffusion models that can be guided to generate samples with additional constraints at test time. We follow Song \etal~\cite{song2020denoising}. After each diffusion steps, we hijack the additional constraints with corresponding noise levels for the next diffusion step. 

More specifically, instead of passing in the network's output $\bm{x}_t$ from the previous time step, we hijack it with $\bm{x}_t \leftarrow \tilde{\bm{x}}_t\bm{m} + \bm{x}_t (1-\bm{m})$, where $\bm{m}$ is the indicator mask of the given condition $\tilde{\bm{x}}_0$. The unspecified constraints in $\tilde{\bm{x}}_0$ are set to 0. $\tilde{\bm{x}}_t$ represents the additional constraint with corresponding noise level, \ie $\sqrt{1-\bar \alpha_t}\tilde{\bm{x}}_0 + \sqrt{\bar{\alpha_t}} \bm{\epsilon}$.


\subsection{ContentNet (Sec3.2)}\label{content}
\label{sec:content}
The goal of ContentNet is to generate high-resolution ($256^2$) realistic HOI images conditioned on the predicted layout and the input object image. We tried two different approaches commonly used in diffusion models~\cite{ldm,nichol2021glide} as backbones for the ContentNet. One way (called ours/AffordDiff-LDM) is to follow Rombach \etal~\cite{rombach2022high}, as described in our main paper, that implements the ContentNet in the latent space where images of size $256^2$ are compressed to 3-dimensional features of size $64^2$ by a fixed pretrained autoencoder. The other way (called ours/AffordDiff-GLIDE) is to follow Nichol \etal~\cite{nichol2021glide} that uses a cascaded diffusion model that first generates images of size $64^2$ and then upsamples them by a factor of $4$.   

\textit{All} of the quantitative results in our main paper, including the user studies and all ablations, are based on Afford-LDM. AffordDiff-GLIDE is better in terms of contact recall ($90.8\%$ vs $87.1\%$) while AffordDiff-LDM is significantly better in terms of FID score ($99.0$ vs $121.6$).   We find that AffordDiff-LDM generates less blurry results and the hand texture appears sharper and more realistic.  In comparison, we find AffordDiff-GLIDE  perceptually preferred because AffordDiff-GLIDE generates more realistic, though blurrier, finger articulations. The qualitative results in the main paper on EPIC-KITCHEN dataset (Fig 1 and Fig4 right in the main paper) show Afford-GLIDE. However, we provide the qualitative comparison of Afford-LDM with baselines in Fig~\ref{fig:more_hoi4d} and Fig~\ref{fig:more_epic} of the appendix.   We further provide a comparison of these two variants in Fig~\ref{fig:glide_ldm} of the appendix. 


\subsection{Constructing Paired Training Data (Sec3.3)}\label{3.3}
 \noindent\textbf{
Cropping Details.} We crop all objects with 80\% squared padding before resizing such that objects (hands) appear in similar (different) sizes. The model learns the priors of their relative scales, \eg, a hand to grasp a kettle appears much smaller than that of a mug (Fig 4).

We show that the proposed method to obtain pixel-aligned pairs of HOI and object-only images is robust and can also be applied to more cluttered images. When there is more than one hand in the HOI image, we randomly select one to remove. We show results of applying our data construction method on the HOI4D (Fig~\ref{fig:more_hoi4d}) and the EPIC-KITCHEN (Fig~\ref{fig:more_epic}) datasets.

\subsection{Baselines Implementation}\label{3.4}
\noindent\textbf{Pix2Pix\cite{pix2pix} (Sec4.1)} We modify the official Pix2Pix implementation\footnote{https://github.com/junyanz/pytorch-CycleGAN-and-pix2pix}. Given the predicted layout and the provided object image, we concatenate them channel-wise and pass them through 6 blocks of ResNet to output HOI images. The discriminator takes in the concatenation the of the object-only image, the splatted layout image, and generated HOI image and learns to discriminate between the real and fake domains.  We tried batchnorm and instancenorm and found that batchnorm generated better results in general but has some black holes if the background statistics deviate from that of the training set.

\noindent\textbf{VAE\cite{kingma2013auto} (Sec4.1)} VAE is notoriously known for being hard to balance for both generation variance and reconstruction quality. We sweep hyperparameters of the KL divergence loss's weights from $1, 1e-1, 1e-2, 1e-3, 1e-4$ and use $1e-3$ as it produces the highest contact recall.

\noindent\textbf{GANHand\cite{ganhand} (Sec4.2)}
GANHand is originally proposed both to predict 3D MANO hands for images of YCB objects~\cite{calli2015benchmarking} and to optimize physical plausibility with respect to the known or reconstructed 3D shapes of YCB objects. We compare our method with their sub-network for grasp prediction from RGB images (blue branch in their original paper, Fig 4). The sub-network takes in the object's identity, the desk plane equation and the object's center in 3D space, in addition to the object image. Since these are not available in the HOI4D dataset, we set them to zeros. We apply an additional reconstruction loss for 3D hand joints, MANO hand parameters and camera parameters.  We finetune the network from the public checkpoints for another 10k iterations. 

\subsection{Scene Integration}\label{3.5}
 We integrate our object-centric HOI synthesis to scene-level affordance prediction. We first detect the objects in the scene and then expand the detected bounding box's size with the same pad ratio (0.8 of the original object size). However, when the scene is crowded, the extended object crops may include other objects thus distracting the layout generation. We instead crop the object with the detected bounding box and pad the cropped object with boundary values. This allows the network to generate hand interaction only for the object of interest.

\subsection{Limitation and Failure Cases}\label{3.6}
Although it is encouraging that the proposed model can perform zero-shot generalization to the EPIC-KITCHEN dataset, the proposed method inherits limited generalization capabilities from general learning-based algorithms. The proposed model will fail when the object image's appearance deviates too much from the training set, \textit{e.g.} for too cluttered scenes, extreme lighting, very large objects (like a fridge) or very small objects (like a pin), \etc. The current model also cannot generate hands entering from the top of the frame or generate hands from a third-person's view due to the bias in the training set. These limitations require training with more diverse data. 
Additionally, the consistency of the hand's appearance and of the extracted hand poses can be further improved.

\section{Qualitative Results}\label{qr}
\begin{figure*}
    \centering
    \includegraphics[width=\linewidth]{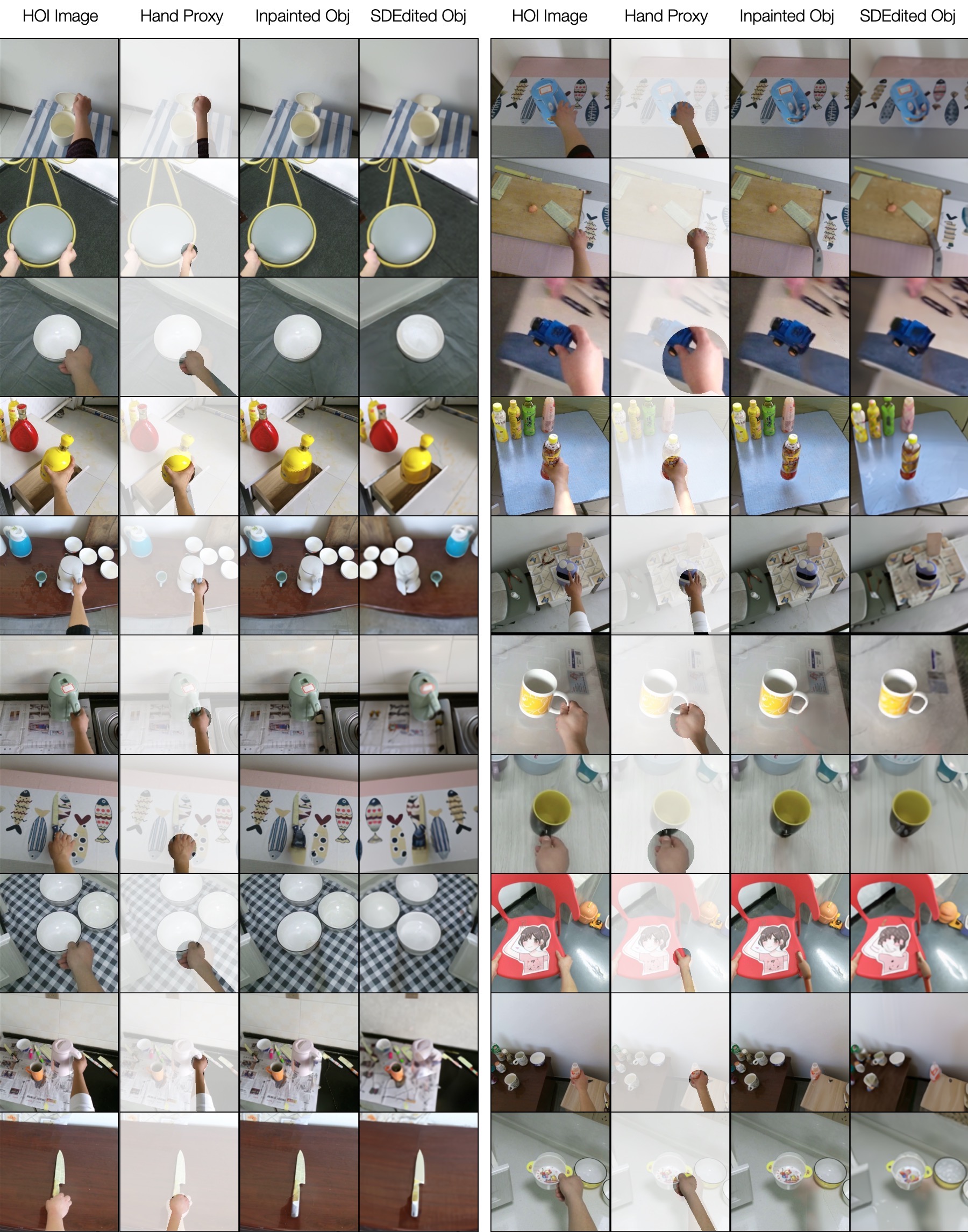}
    \caption{Visualizing more examples of the constructed paired training data. We train all the models with a mixture of inpainted and SDEdited object images.}
    \label{fig:more_hoi4d}
\end{figure*}

Fig~\ref{fig:more_hoi4d} shows more examples of the constructed paired training data. We train all the models with a uniform mixture of inpainted and SDEdited object images.

\begin{figure*}
    \centering
    \includegraphics[width=\linewidth]{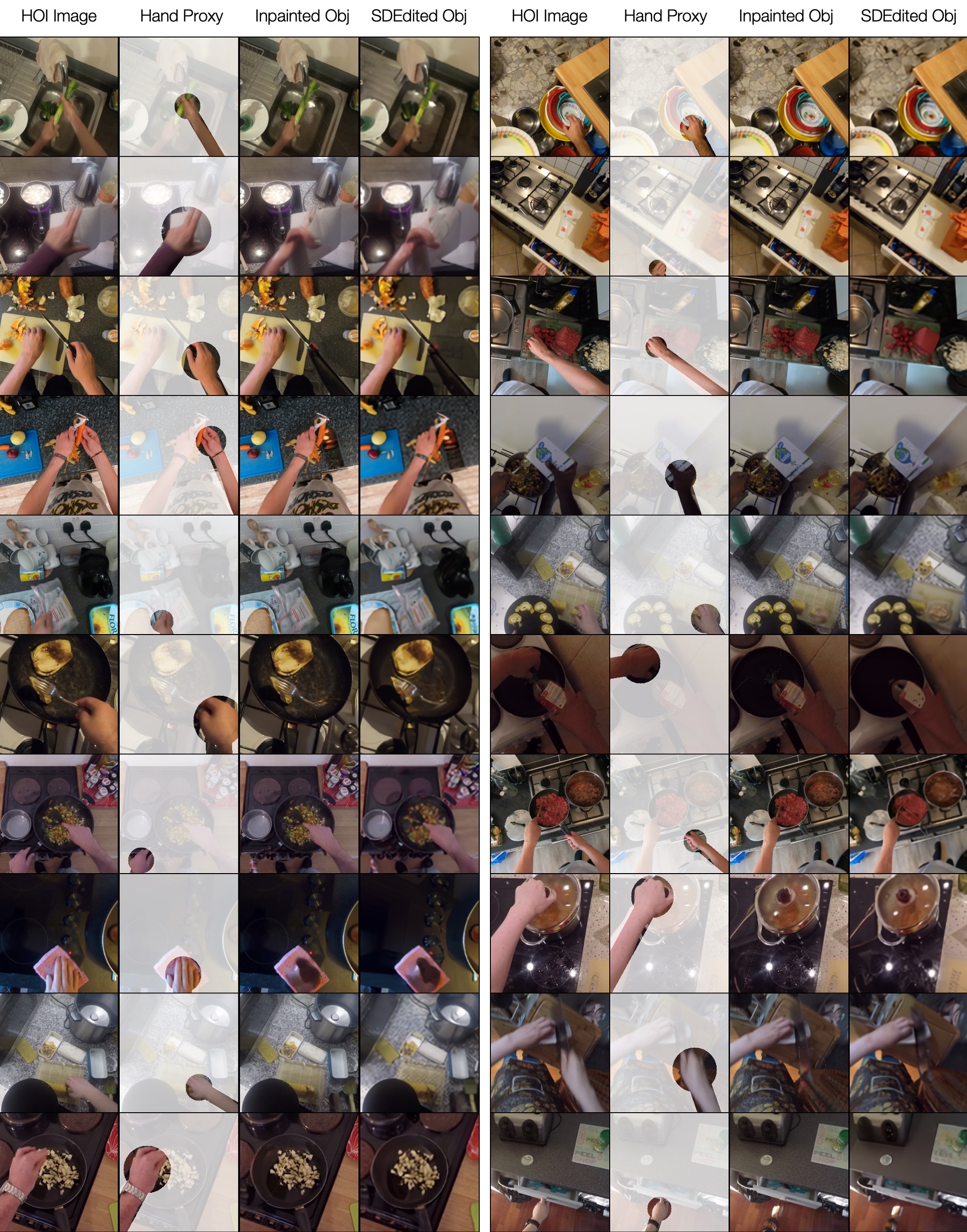}
    \caption{Visualizing the proposed paired data construction applied to EPIC-KITCHEN. }
    \label{fig:more_epic}
\end{figure*}

Fig~\ref{fig:more_epic} shows that the proposed paired data construction is robust and can be applied to the EPIC-KITCHEN dataset.

\begin{figure*}
    \centering
    \includegraphics[width=\linewidth]{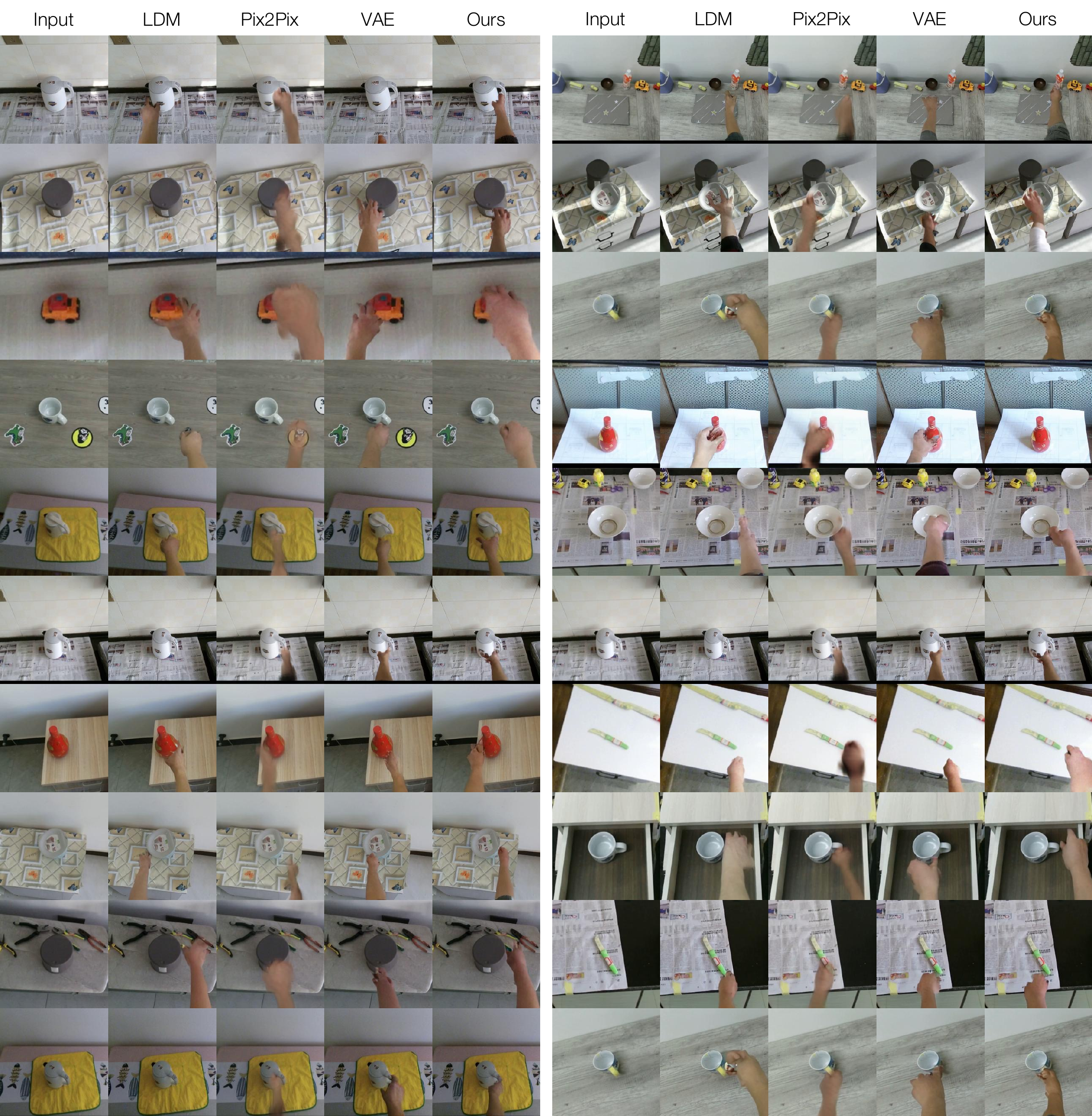}
    \caption{Visualizing more comparisons of the generated HOI images from the proposed method and other image synthesis baselines \cite{ldm,pix2pix,kingma2013auto} on the HOI4D dataset. }
    \label{fig:image_hoi4d}
\end{figure*}

Fig~\ref{fig:image_hoi4d} shows more comparisons of the generated HOI images by the proposed method (LDM-version as reported in tables) and other image synthesis baselines~\cite{ldm,pix2pix,kingma2013auto} on the HOI4D dataset.

\begin{figure*}
    \centering
    \includegraphics[width=\linewidth]{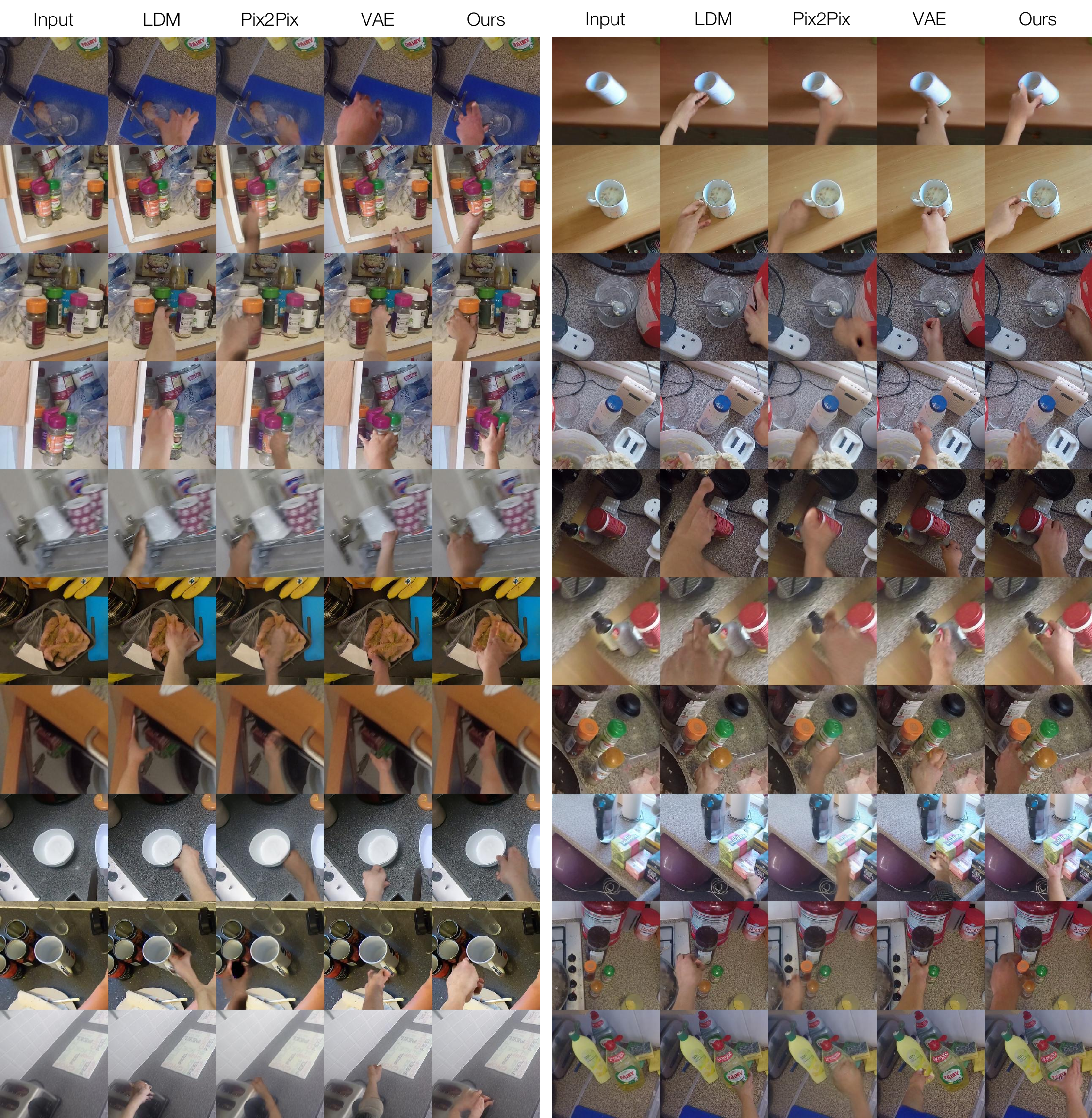}
    \caption{Visualizing more comparisons of the generated HOI images from the proposed method and other image synthesis baselines \cite{ldm, pix2pix,kingma2013auto} on the EPIC-KITCHEN dataset. }
    \label{fig:image_epic}
\end{figure*}

Fig~\ref{fig:image_epic} shows more comparisons of the generated HOI images by the proposed method (LDM-version as reported in tables) and other image synthesis baselines~\cite{ldm,pix2pix,kingma2013auto} on EPIC-KITCHEN dataset.

\begin{figure*}
    \centering
    \includegraphics[width=\linewidth]{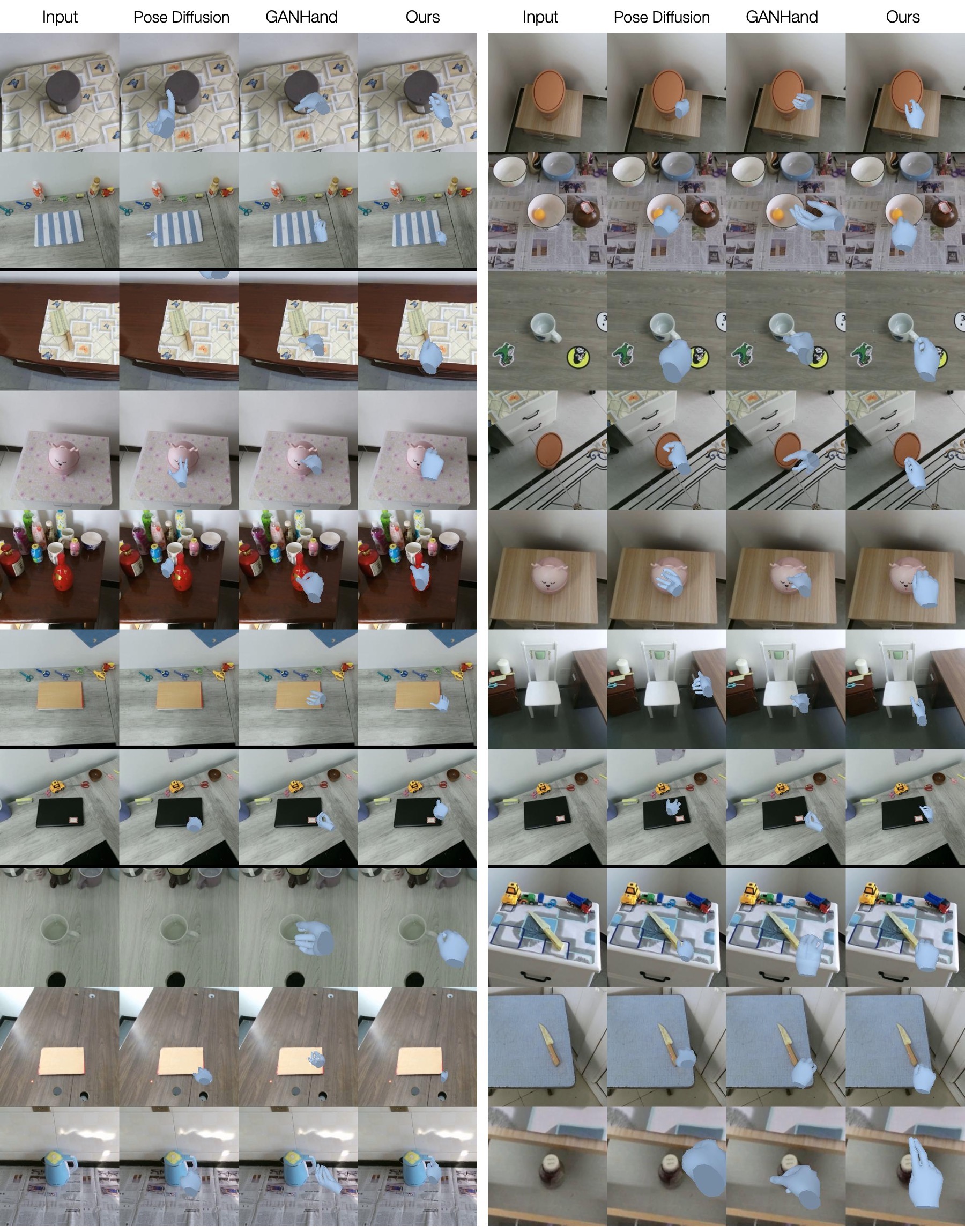}
    \caption{Visualizing more comparisons of the extracted 3D hand pose from the proposed method and other 3D affordance baselines \cite{ganhand,ldm} on the HOI4D dataset. }
    \label{fig:pose_hoi4d}
\end{figure*}

Fig~\ref{fig:pose_hoi4d} shows more comparisons of the extracted 3D hand pose obtained by the proposed method and other 3D affordance baselines~\cite{ganhand,ldm} on the HOI4D dataset.

\begin{figure*}
    \centering
    \includegraphics[width=\linewidth]{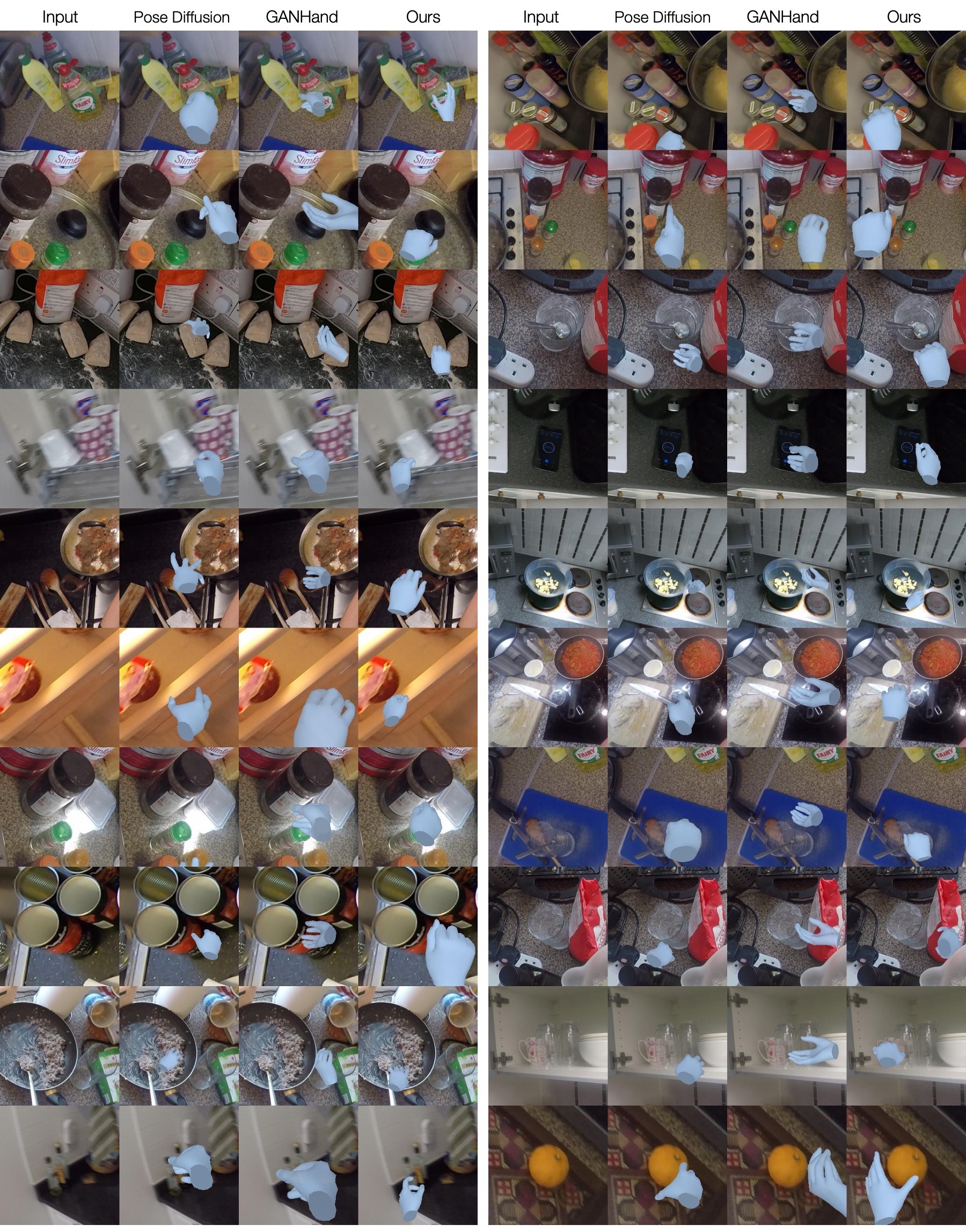}
    \caption{Visualizing more comparisons of the extracted 3D hand pose from the proposed method and other 3D affordance baselines on the EPIC-KITCHEN dataset. }
    \label{fig:pose_epic}
\end{figure*}

Fig~\ref{fig:pose_epic} shows more comparisons of the extracted 3D hand pose obtained by the proposed method and other 3D affordance baselines~\cite{ganhand,ldm} on the EPIC-KITCHEN dataset.

\begin{figure*}
    \centering
    \includegraphics[width=\linewidth]{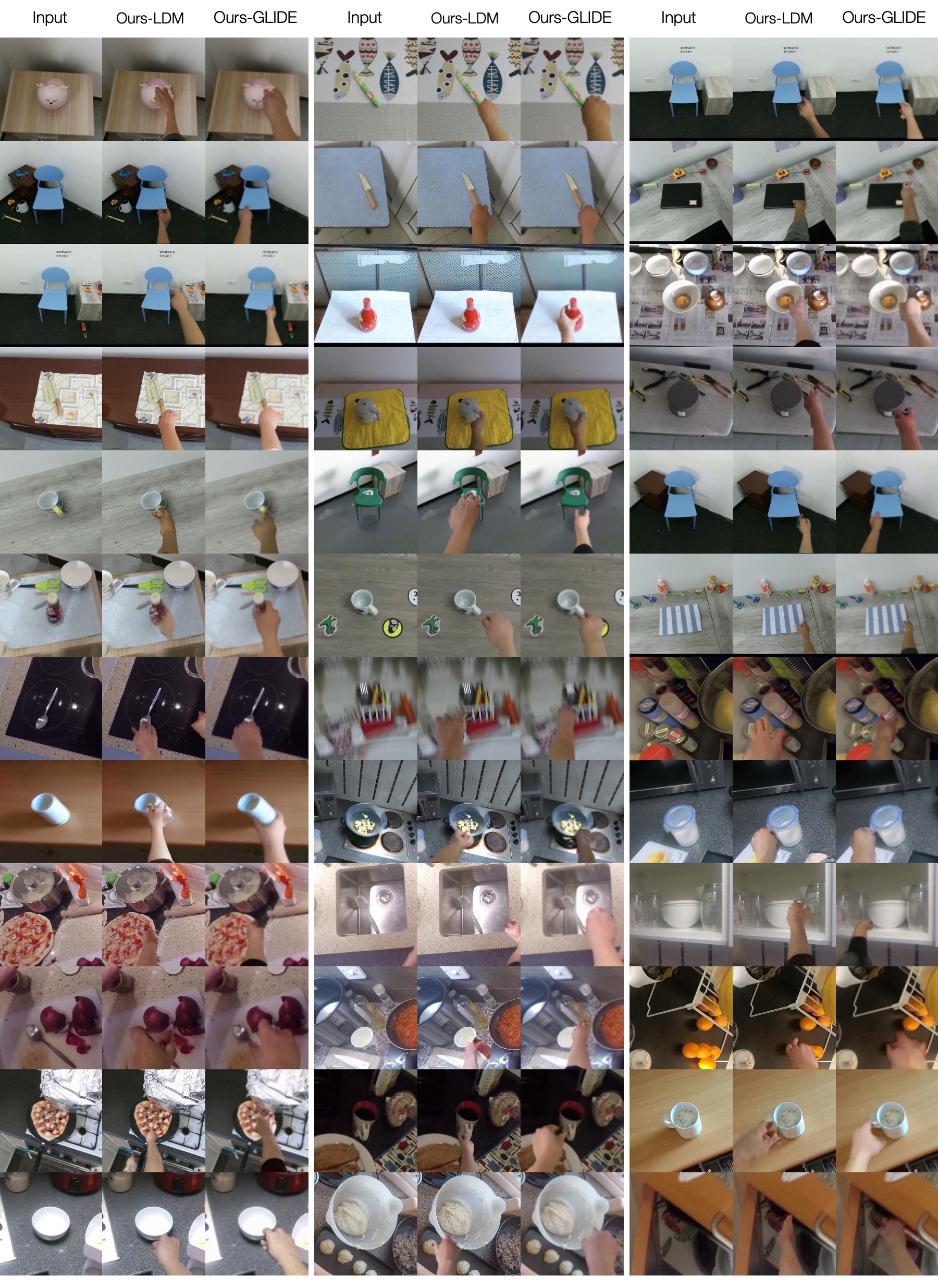}
    \caption{Visualizing the ablation of ContentNet for its LDM-based and GLIDE-based implementations (Sec~\ref{sec:content}).}
    \label{fig:glide_ldm}
\end{figure*}

Fig~\ref{fig:glide_ldm} shows an ablation study on comparison of the LDM and GLIDE version of our model on HOI4D and EPIC-KITCHEN datasets. 

\begin{figure*}
    \centering
    \includegraphics[width=\linewidth]{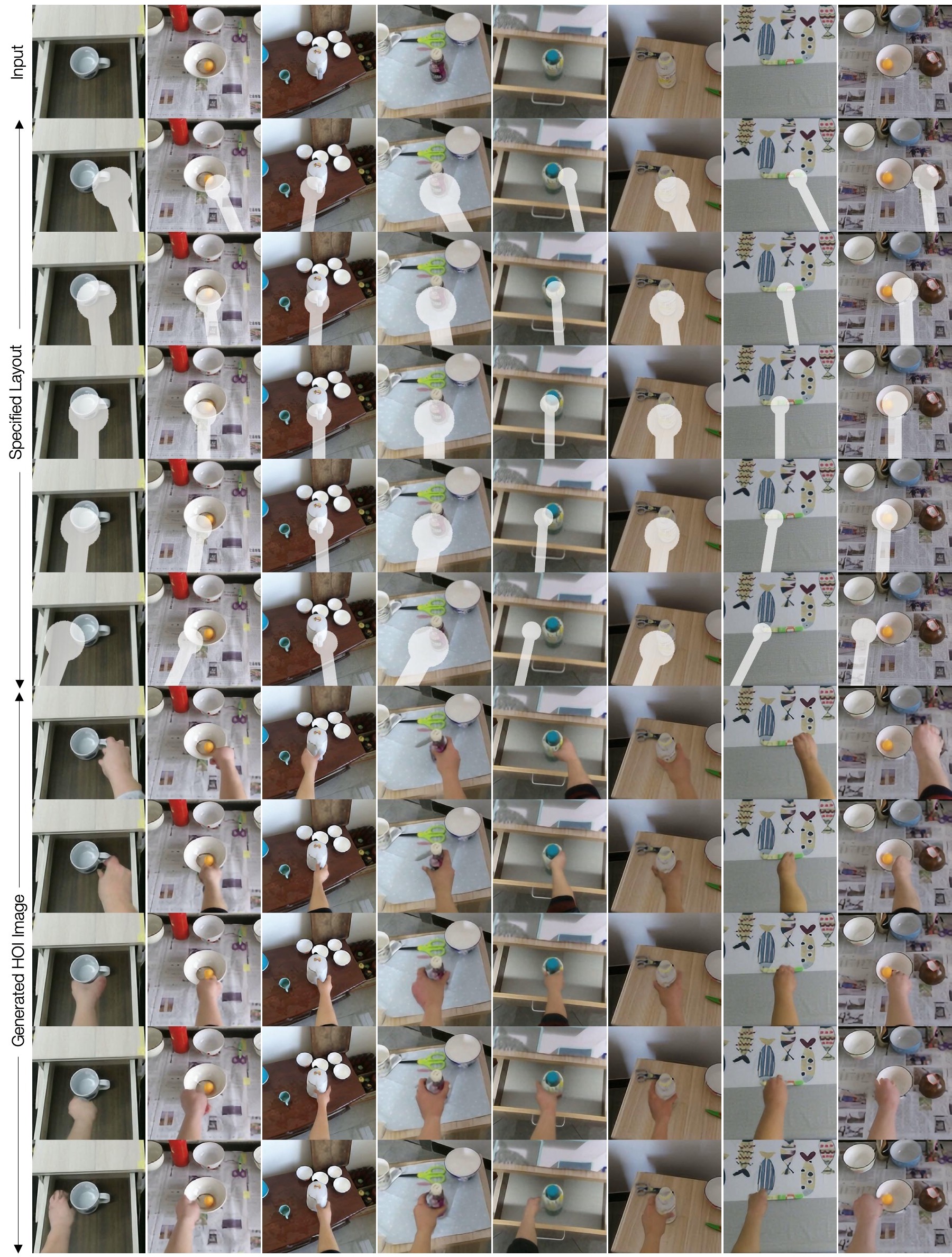}
    \caption{Visualizing more layout editing results. }
    \label{fig:more_inter}
\end{figure*}

Fig~\ref{fig:more_inter} shows more layout editing results. 

\begin{figure*}
    \centering
    \includegraphics[width=\linewidth]{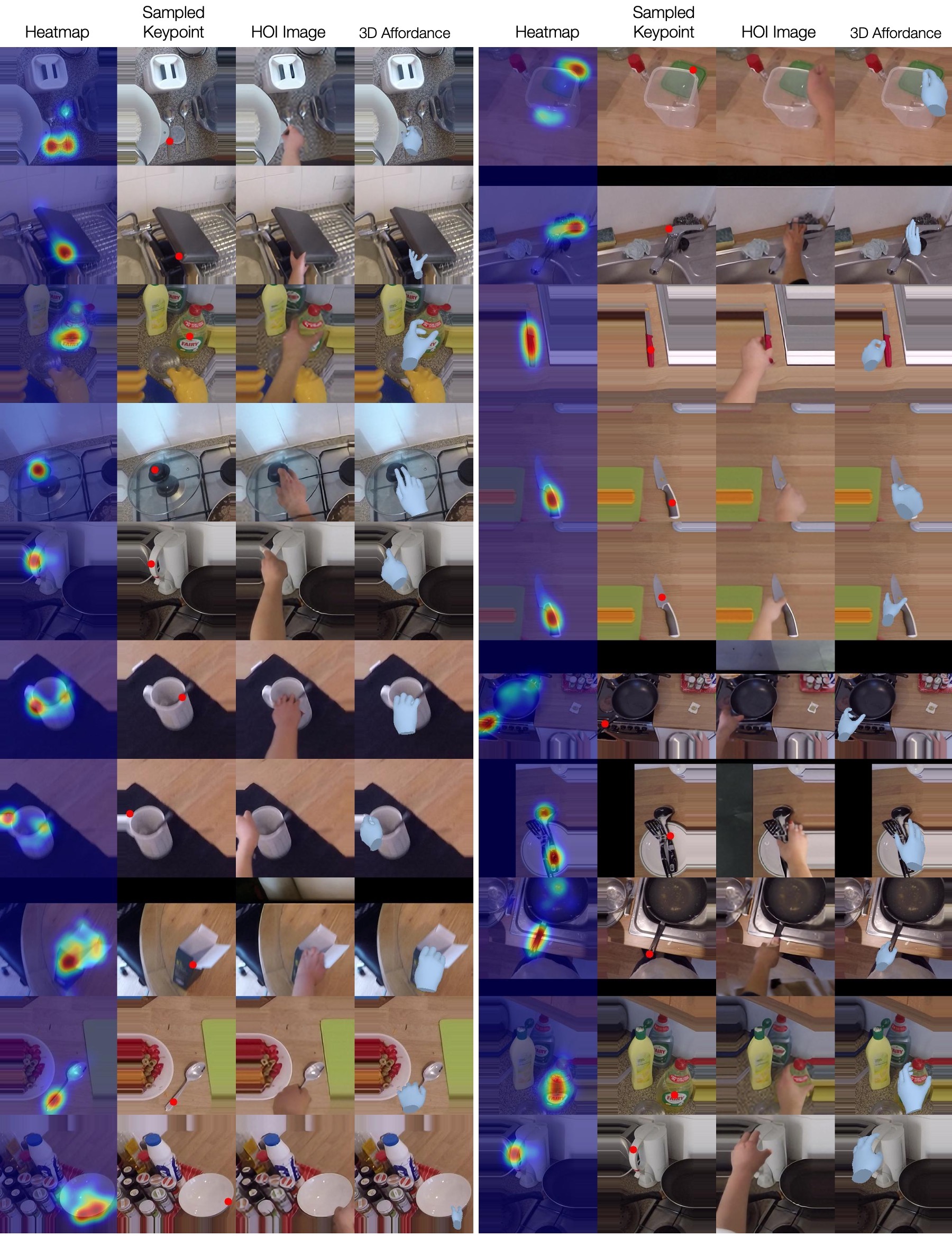}
    \caption{Visualizing more results of heatmap-guided synthesis.  }
    \label{fig:more_heatmap}
\end{figure*}

Fig~\ref{fig:more_heatmap} shows more results of heatmap-guided synthesis. 

\clearpage

\begin{figure*}
    \centering
    \includegraphics[width=\linewidth]{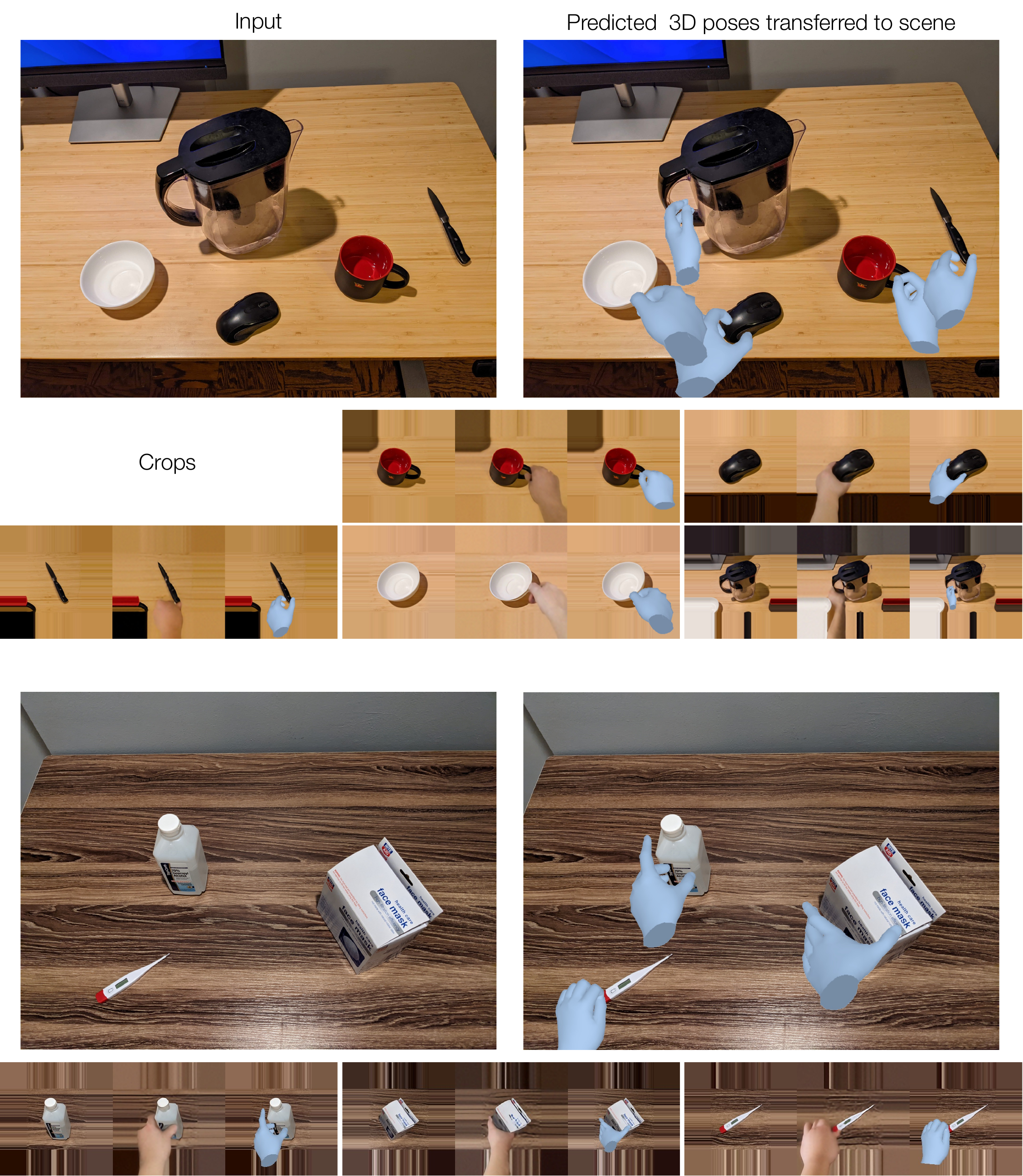}
    \caption{Visualizing more scene integration results with the individual prediction from crops. }
    \label{fig:more_scene}
\end{figure*}

\end{document}